\documentclass[10pt,twocolumn,letterpaper]{article}

\usepackage{cvpr}              
\usepackage{verbatim}
\usepackage{multirow}
\usepackage{pifont}
\usepackage{colortbl}  
\usepackage{xcolor}    
\makeatletter
\renewcommand\subsubsection{\@startsection{subsubsection}{3}{0mm}{0.1\baselineskip}{-1em}{\bfseries}}
\makeatother

\definecolor{cvprblue}{rgb}{0.21,0.49,0.74}
\usepackage[pagebackref,breaklinks,colorlinks,allcolors=cvprblue]{hyperref}

\def\paperID{6187} 
\def\confName{CVPR}
\def\confYear{2025}

\title{RC-AutoCalib: An End-to-End Radar-Camera Automatic Calibration Network}

\author{
Van-Tin Luu$^{1}$, Yon-Lin Cai$^{1}$, Vu-Hoang Tran$^{2}$, Wei-Chen Chiu$^{1}$, Yi-Ting Chen$^{1}$, Ching-Chun Huang$^{1}$\thanks{Corresponding author}\\
$^{1}$National Yang Ming Chiao Tung University, Taiwan\\
$^{2}$Ho Chi Minh City University of Technology and Education, Vietnam\\
{\tt\small \{tinery.ee12, yukitaka.10, walon, ychen, chingchun\}@nycu.edu.tw}\\
{\tt\small hoangtv@hcmute.edu.vn}
}

\begin{document}
\maketitle
\begin{abstract}
This paper presents a groundbreaking approach - the first online automatic geometric calibration method for radar and camera systems. Given the significant data sparsity and measurement uncertainty in radar height data, achieving automatic calibration during system operation has long been a challenge. To address the sparsity issue, we propose a Dual-Perspective representation that gathers features from both frontal and bird’s-eye views. The frontal view contains rich but sensitive height information, whereas the bird’s-eye view provides robust features against height uncertainty. We thereby propose a novel Selective Fusion Mechanism to identify and fuse reliable features from both perspectives, reducing the effect of height uncertainty. Moreover, for each view, we incorporate a Multi-Modal Cross-Attention Mechanism to explicitly find location correspondences through cross-modal matching. During the training phase, we also design a Noise-Resistant Matcher to provide better supervision and enhance the robustness of the matching mechanism against sparsity and height uncertainty. Our experimental results, tested on the nuScenes dataset, demonstrate that our method significantly outperforms previous radar-camera auto-calibration methods, as well as existing state-of-the-art LiDAR-camera calibration techniques, establishing a new benchmark for future research. The code is available at https://github.com/nycu-acm/RC-AutoCalib
\end{abstract}    
\vspace{-0.4cm}
\section{Introduction}
\label{sec:intro}
\begin{figure}[t]
    \centering
    \includegraphics[width=1\linewidth]{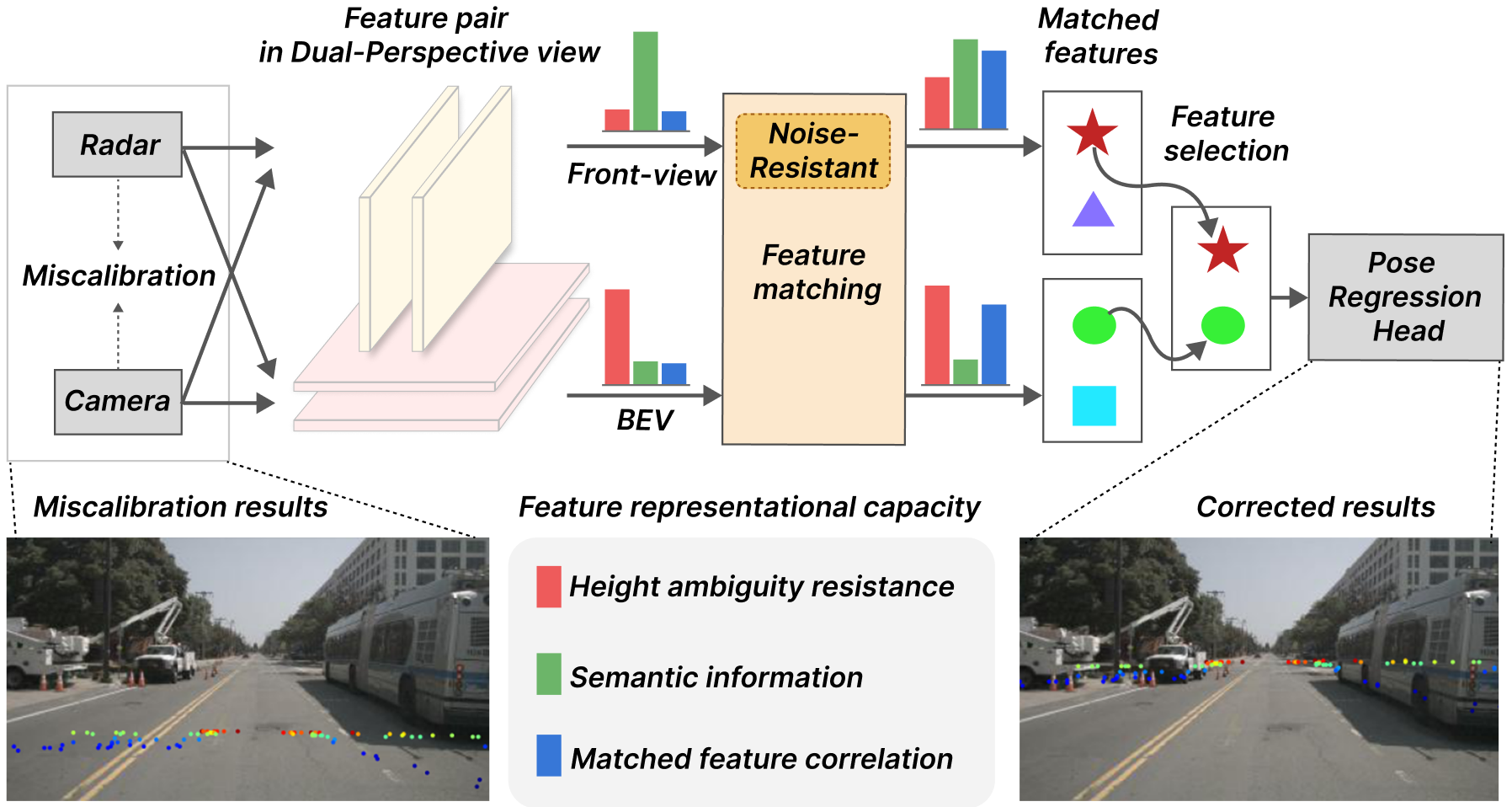}
    \caption{ An overview of the proposed RC-AutoCalib method. The approach takes input from radar-camera miscalibration, representing it as feature pairs in Dual-perspective view. These feature representations are then enhanced through feature matching block, from which reliable features are selected to predict the rotation vector and translation.}
    \vspace{-0.5cm}
    \label{fig:highlevel}
\end{figure}

Radars and cameras are increasingly favored in advanced driver-assistance systems (ADAS) due to their cost-effectiveness and robust performance in diverse weather conditions. A critical research area in these systems involves data fusion and multi-modal calibration to ensure reliable functioning in real-world settings \cite{long2021radar,wu2022sparse,Zhuang_2021_ICCV,li2022bevdepthacquisitionreliabledepth,yang2020radarnet}. Conventional calibration techniques for 3D radar and cameras primarily focus on offline methods, which often rely on specialized calibration targets like checkerboards or corner reflectors, and are generally limited to the radar measurement plane \cite{sugimoto2004obstacle,wang2011integrating,kim2014data,kim2018radar,kim2017comparative,el2015radar,domhof2019extrinsic}. These methods, while effective, require substantial time and manual effort, and they do not account for sensor displacements that can occur under normal operating conditions. This limitation underscores the necessity for online auto-calibration, which can dynamically adjust to changes over time.

Online auto-calibration methods eliminate the need for calibration targets, focusing on matching natural features collected by radar and camera sensors. While this approach offers greater flexibility in real-world scenarios, exploration in this area remains limited, with no established benchmarks using publicly available datasets to date. Only the approach by Schöller et al. \cite{scholler2019targetless} utilizes deep learning to address the problem of online auto-calibration for radar and camera. However, their focus remains exclusively on rotational calibration, without addressing translational calibration.

In contrast, online auto-calibration methods for LiDAR and camera \cite{wang2022fusionnet,lv2021lccnet,schneider2017regnet,zhao2021calibdnn,iyer2018calibnet,shi2020calibrcnn,lv2021cfnet,wu2021netcalib,wu2021way,zhu2023calibdepth,jing2022dxq,Liu2021SemAlignAC,yuan2021pixel} have been extensively explored by researchers and have demonstrated powerful capabilities. Most of these methods share a common concept of using RGB images and mis-calibrated LiDAR data as input, with the overall process divided into feature extraction, feature matching, and parameter regression. Instead of directly extracting features from RGB images \cite{lv2021lccnet,wang2022fusionnet,zhao2021calibdnn,shi2020calibrcnn}, some methods have modified the representation of RGB images. For instance, \cite{Liu2021SemAlignAC,yuan2021pixel} implicitly extract features in the form of semantics and edges, while \cite{zhu2023calibdepth,wu2021netcalib} transform RGB images into depth maps to achieve a unified representation consistent with 3D data. Overall, these methods project 3D data onto the frontal view to fuse with the camera information for further processing.
\begin{figure}[t]
  \centering
  \begin{subfigure}{0.61\linewidth}
    \includegraphics[width=\linewidth]{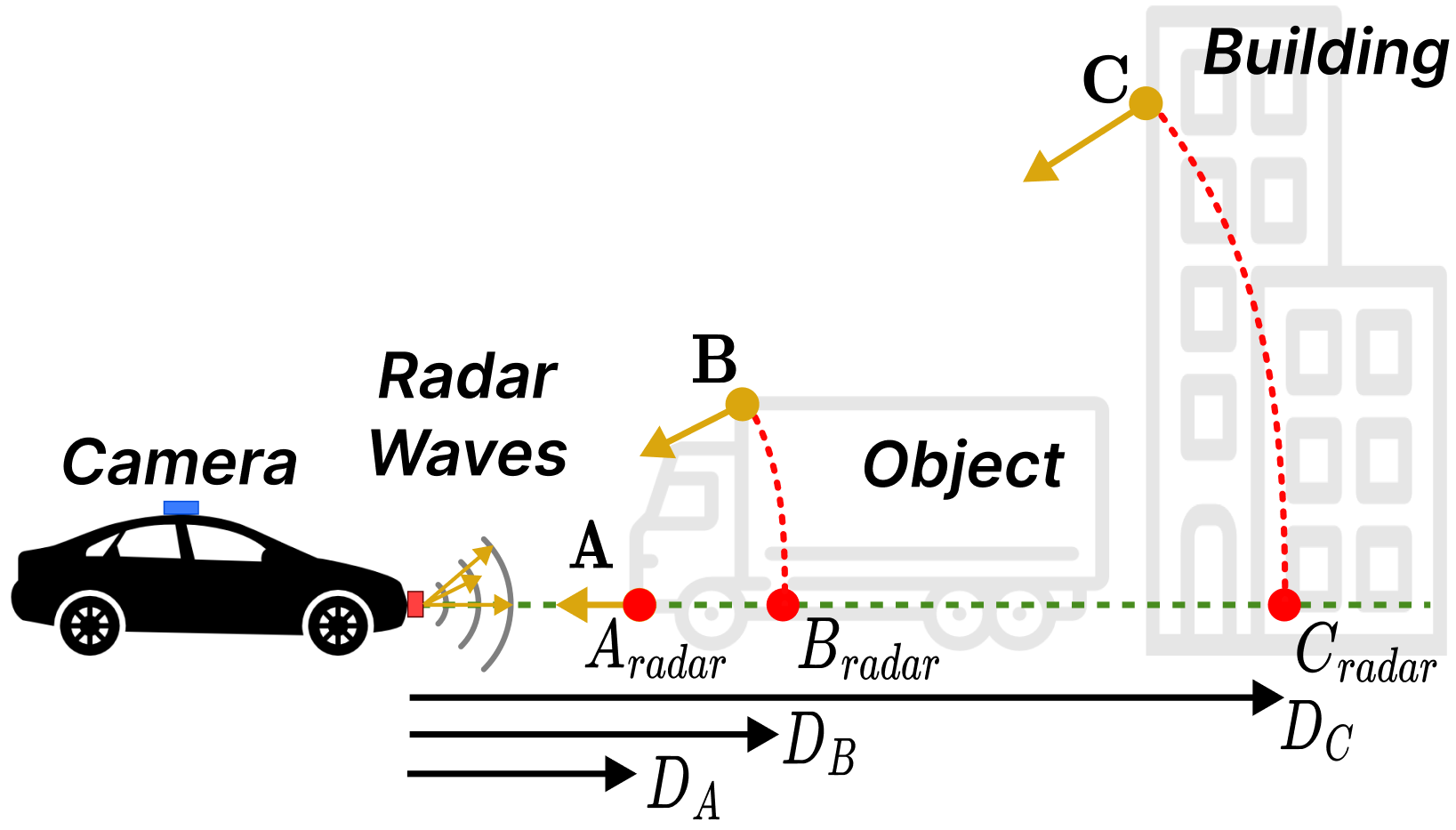}
    \caption{Elevation Ambiguity in Radar Point Clouds~\cite{long2021radar}.}
    \vspace{0.2cm}
    \label{fig:short-a}
  \end{subfigure}
   \hspace{0.001\linewidth}
  \begin{subfigure}{0.36\linewidth}
    \includegraphics[width=\linewidth]{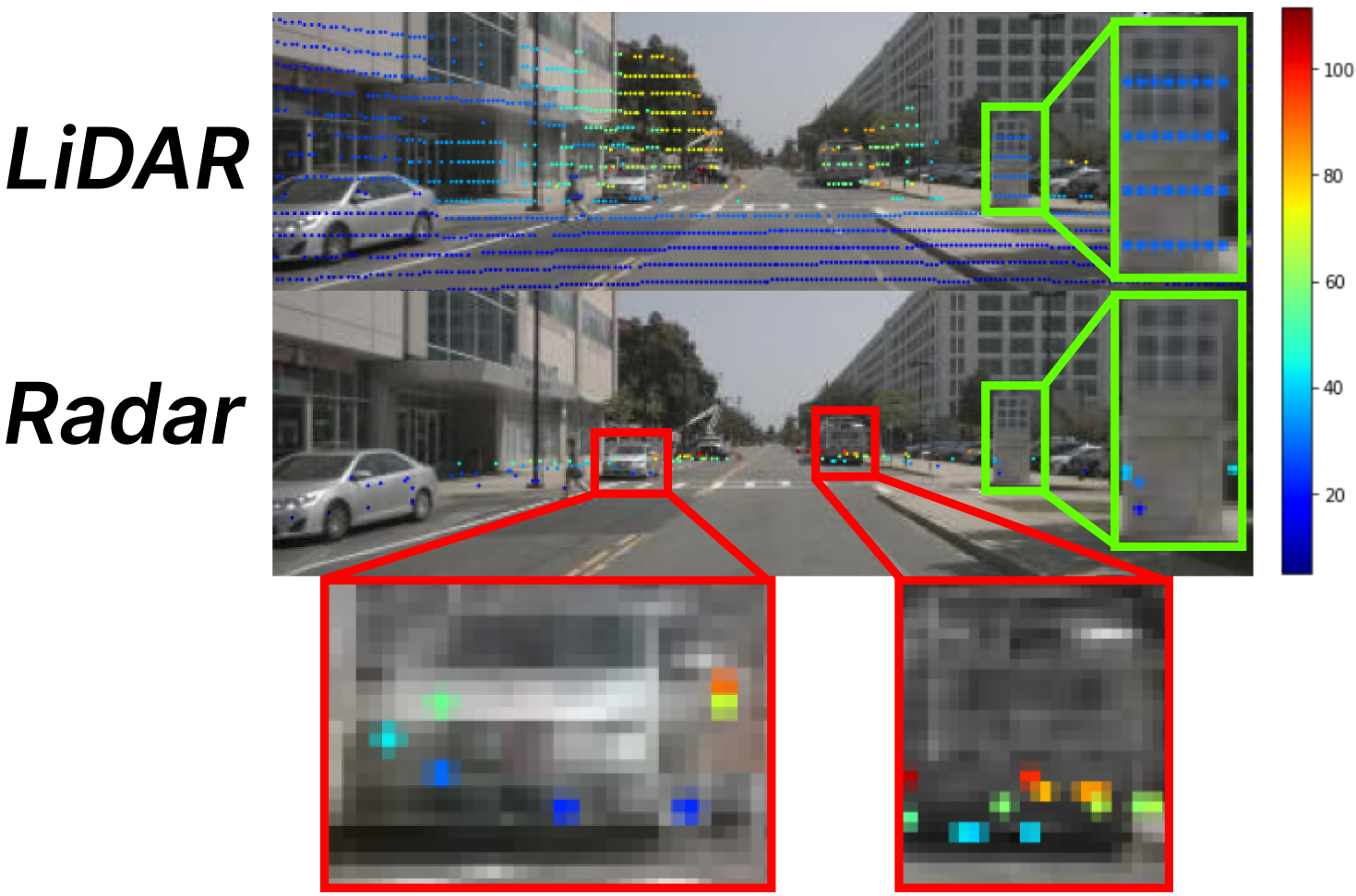}
    \caption{Noisy depth measurements and lack of object structures.}
    \label{fig:short-b}
  \end{subfigure}
  \caption{Challenges of 3D Millimeter-Wave Radar. (a) The green dashed line represents the height plane the radar focuses on. Points $\mathbf{A}$, $\mathbf{B}$, and $\mathbf{C}$ denote actual reflection positions, whereas $\mathbf{A}_{\text{radar}}$, $\mathbf{B}_{\text{radar}}$, and $\mathbf{C}_{\text{radar}}$ are the positions recorded by the radar. $\mathbf{D}_{\text{A}}$, $\mathbf{D}_{\text{B}}$, and $\mathbf{D}_{\text{C}}$ represent the recorded and noisy radar depths. (b) The top image shows a ``LiDAR'' depth map projected onto the camera plane, while the bottom image displays a ``radar'' depth map projected similarly. The red box highlights the issue where depths of points on the same object should be similar, yet significant variations are evident, indicating the presence of noise. Moreover, the green box shows a structural comparison: the ``LiDAR'' point cloud distinctly outlines the object’s contour, while the ``radar'' point cloud fails to convey structural information.}
  \vspace{-0.3cm}

  \label{fig:challenge}
\end{figure}
The above-mentioned LiDAR-camera methods provide a reference and comparison for developing radar-camera auto-calibration. However, we find that relying solely on a single viewpoint for radar-camera auto-calibration makes achieving high accuracy challenging. As depicted in \cref{fig:challenge}, the frontal depth map often contains noisy point cloud data due to uncertainty from the lack of radar height information. Additionally, radar data is inherently sparse and lacks object structures. When projecting radar point clouds onto the frontal view to form the depth map, the projected points tend to overlap and become even sparser. To mitigate these challenges, we have introduced a Dual-Perspective representation that leverages attention-based selection to extract more reliable features.


Furthermore, feature matching between radar and camera is a critical component for auto-calibration. Some methods \cite{zhu2023calibdepth,shi2020calibrcnn,wu2021netcalib} rely on concatenation followed by several convolutional layers to facilitate feature matching, while others \cite{lv2021lccnet} use cost volume to represent the correlation between the two sensors. However, traditional approaches depend solely on implicit supervision from the final calibration loss to guide the matching process. This lack of explicit identification of matched local pairs between sensors renders the calibration process indistinct. To address this, we have developed a Noise-Resistant Matcher that provides direct supervision for feature matching and correspondence finding.

Accordingly, as illustrated in \cref{fig:highlevel}, we propose RC-AutoCalib, an end-to-end network for automatic 3D radar and camera calibration, addressing the challenges of sparse and noisy radar data. To counteract these issues, we enhance a Dual-Perspective representation that integrates features from both the frontal view and the bird's eye view (BEV). The frontal view is prone to noise due to missing height information in radar point clouds, while the BEV provides more stable features, unaffected by this limitation. Our model includes a Selective Fusion Mechanism to discern and utilize beneficial features from each perspective. Additionally, we incorporate a Multi-Modal Cross-Attention Mechanism to focus on relevant areas in sparse radar point clouds. To improve calibration accuracy, we introduce Explicit Feature Matching Supervision with a Noise-Resistant Matcher, which helps the model identify and learn from correspondence points between radar and camera, filtering out noise in the process. Our results on the nuScenes dataset demonstrate significant improvements over existing LiDAR-camera calibration methods, as well as previous radar-camera auto-calibration approaches, setting a new benchmark for future research. In summary, our contributions are:
\begin{itemize}
\renewcommand{\labelitemi}{$\circ$} 
\item We introduce RC-AutoCalib, an end-to-end network for calibrating 3D radar and cameras, featuring a Dual-Perspective representation that counters the height information limitations of 3D radar data. This network includes a novel Selective Fusion Mechanism to optimally integrate features from both the frontal view and BEV perspectives.
\item We develop a feature-matching module incorporating a Multi-Modal Cross-Attention Mechanism to enhance the utilization of radar point clouds. This module integrates  a Noise-Resistant Matcher to provide Explicit Feature Matching Supervision. Thereby, RC-AutoCalib can effective filter out noise caused by height inaccuracies and enable robust learning of radar-image correspondences for calibration.

\item Our approach demonstrates superior experimental results on the nuScenes dataset compared to existing LiDAR-camera calibration methods, establishing a new benchmark for future research.
\end{itemize}

\section{Related Works}
\label{sec:related}

\subsection{Offline Calibration}
Offline calibration methods primarily depend on specific calibration targets and cannot address real-time errors. These methods are tailored for fixed environments and necessitate substantial manual effort to achieve precision, rendering them unsuitable for dynamic conditions and generally reserved for controlled settings. Early radar-camera calibration techniques focused on merging radar signals with camera data through homography projection that maps points from the radar's horizontal plane to the camera image plane. Due to inherent noise in radar sensors, these early methods often required specialized trihedral reflectors to establish accurate correspondences \cite{sugimoto2004obstacle,wang2011integrating,kim2014data,kim2018radar}. However, the radar's limitation in accurately measuring the elevation of distant targets indicated that reflectors had to be positioned precisely on the radar's horizontal plane \cite{sugimoto2004obstacle}. More recent radar calibration algorithms aim to minimize ``reprojection error'' to better synchronize object detection across both sensor fields of view, using techniques like estimating radar-to-camera transformations via reprojection error \cite{kim2017comparative}, or intersecting back-projected camera rays with 3D ``arcs'' that conform to radar measurements to determine necessary transformations \cite{el2015radar}. Despite improvements, these methods still rely on specific targets and manual input efforts.

\subsection{Online Calibration}
Online methods primarily extract features from natural scenes for calibration, offering greater flexibility and adaptability to various scenarios. The rapid development of deep learning has demonstrated neural networks' powerful feature extraction capabilities. However, due to the aforementioned challenges associated with radar, online calibration methods for radar and cameras are less prevalent. In this paper, we focus on developing an end-to-end architecture for the online auto-calibration of radar and cameras, leveraging robust benchmarks established by LiDAR and camera calibration methods.

\subsubsection*{LiDAR and Camera.} Li et al. \cite{li2023automatic} categorized targetless calibration methods into information theory-based, feature-based, ego-motion-based, and learning-based approaches. Pandey et al. \cite{pandey2012automatic} used mutual information between point cloud intensities and image grayscale values. Taylor and Nieto \cite{taylor2015motion} utilized sensor ego-motion on moving vehicles to estimate extrinsic parameters. Levinson and Thrun \cite{levinson2013automatic} as well as Yuan et al. \cite{yuan2021pixel} optimized depth-discontinuous and depth-continuous edge features, respectively. Regnet \cite{schneider2017regnet} and CalibNet \cite{iyer2018calibnet} employed deep learning to match features and regress calibration parameters. CalibRCNN \cite{shi2020calibrcnn} combined CNN with LSTM \cite{sak2014long} and added pose constraints for accuracy. LCCNet \cite{lv2021lccnet} used cost volume for feature correlation. Despite achieving positive results, these methods do not explicitly learn the correspondence between point clouds and images. In contrast, in this paper, we introduce Explicit Feature Matching Supervision to guide the model in learning the correspondence relationship between point clouds and images more effectively.

\begin{figure*}[t]
    \centering
    \includegraphics[width=.85\linewidth]{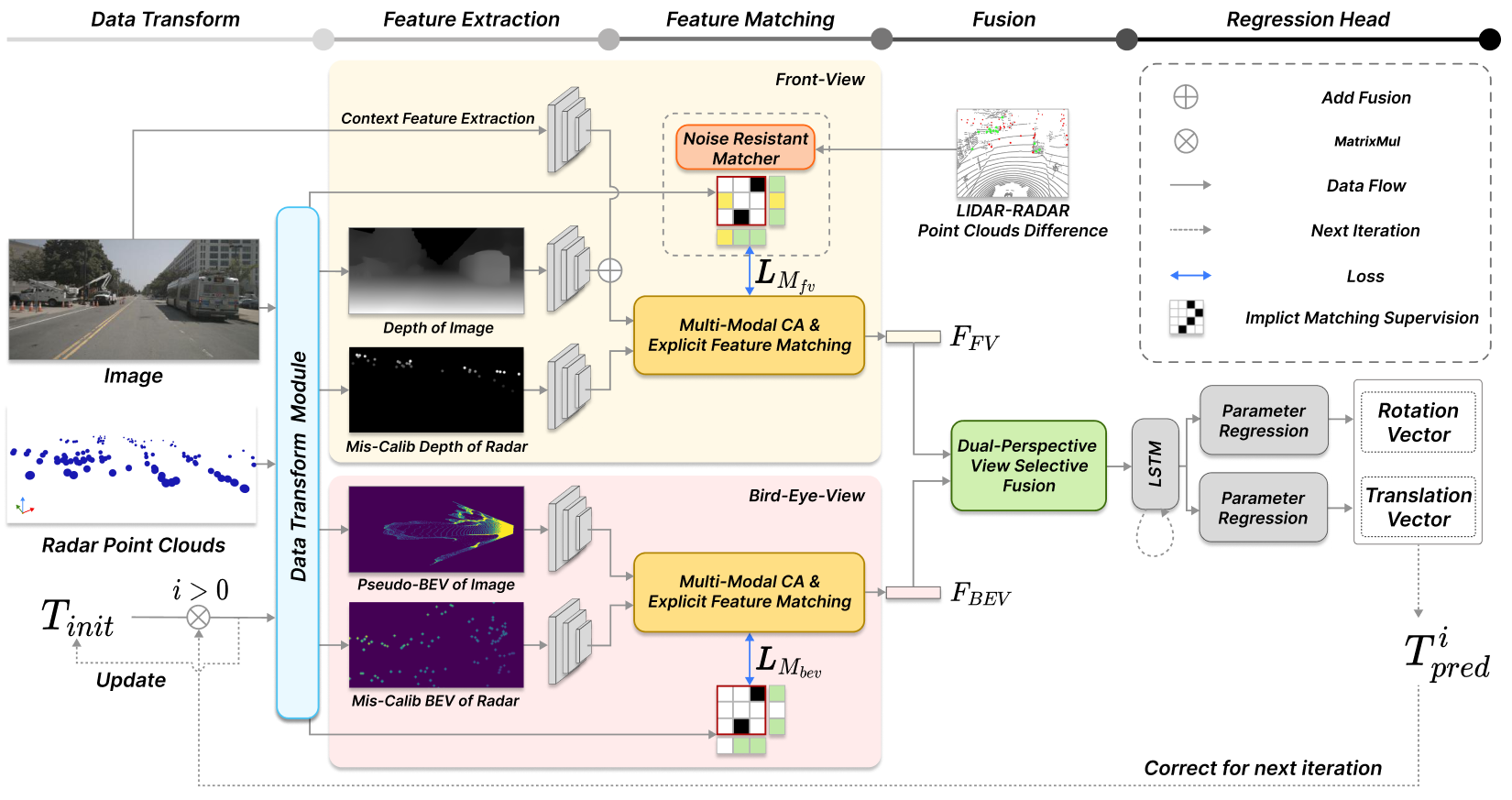}
    \caption{ Our system flow for iterative online auto-calibration starts with the input image, point cloud, and initial calibration parameters \( T_{init} \), which first pass through the Data Transform module (\cref{sec:data transform}). Here, we obtain the estimated image depth map and miscalibrated radar depth map from the frontal view (FV) perspective, along with the pseudo-BEV image and miscalibrated BEV radar projection. These outputs are then processed in the Feature Extraction module (\cref{sec:feature extraction}), where features from both FV and BEV perspectives undergo Feature Matching (\cref{sec:feature matching}) between the image and radar data. Subsequently, after Feature Matching and Fusion (\cref{sec:selective fusion}), the Regression Head (\cref{sec:regression head}) generates the rotation and translation vectors that form the transformation matrix, $\hat{T}_{pred}^{i}$, to refine calibration. Finally, $\hat{T}_{pred}^{i}$ is fed back to \(T_{init}\) to update the calibration parameters for the next $i$-th iteration.}
    \vspace{-0.3cm}
    \label{fig:overall}
\end{figure*}

\subsubsection*{Radar and Camera.} Per{\v{s}}i{'c} et al. \cite{pervsic2021online} proposed an online calibration method based on detecting and tracking moving objects, focusing on rotational calibration. Sch{\"o}ller et al. \cite{scholler2019targetless} used deep learning to learn rotational calibration matrices but did not address translational calibration. Additionally, their methods utilize stationary traffic radars fixed on highway positions, differing from ours that employ vehicle-mounted 3D radars moving with the car. Wisec et al. \cite{wise2021continuous} developed a targetless calibration method for 3D radar and cameras, using radar velocity information and motion-based camera pose measurements, solved with nonlinear optimization. Later, the same research team extended their work \cite{wise2021continuous} to include radar ego-velocity estimates and unscaled camera pose measurements in \cite{wise2023spatiotemporal} for a more complete spatiotemporal calibration. However, these methods overly rely on radar speed measurements, making them less robust to noise. Additionally, they do not leverage the power of deep learning and fail to explicitly establish the correspondence between radar and images.

\section{Methods}

The overall pipeline of the RC-AutoCalib method, depicted in \cref{fig:overall}, begins with RGB images and radar point clouds as inputs. These inputs pass through the Data Transform module, yielding the frontal view estimated depth map, frontal view miscalibrated radar depth map, pseudo-BEV image, and miscalibrated radar BEV. Subsequent processing occurs in the Feature Extraction module, where features are extracted. These features are then analyzed in the Feature Matching module to enhance understanding of the correlation between feature pairs. This module incorporates a Multi-Modal Cross-Attention Mechanism, Explicit Feature Matching Supervision, and a Noise-Resistant Matcher. Following this, the Selective Fusion Mechanism aggregates the features, and the system performs parameter regression to predict rotation and translation vectors necessary for auto-calibration. Detailed descriptions are provided below.



\subsection{Data Transform module}
\label{sec:data transform}

To address issues of uncertainty caused by elevation ambiguity and the sparsity of radar data, we propose a Dual-Perspective feature representation. This approach projects two types of 3D data (i.e., the image plus its depth map and the radar point clouds) onto two different perspectives: bird's-eye view (BEV) and frontal view (FV). The BEV provides a domain where radar data is less impacted by uncertain height and offers more information about the geometry of the scene. Simultaneously, the FV retains rich semantic information, preserving important contextual details.




\subsubsection*{Radar data.} Given a random initialized or roughly-estimated extrinsic transformation \( T_{init} \) \cite{wang2022fusionnet,lv2021lccnet,zhao2021calibdnn,iyer2018calibnet}, consisting of a rotation matrix \( R_{init} \) and a translation vector \( t_{init} \), we transform a 3D radar point \( {P}_r = (X_{r}, Y_{r}, Z_{r})\) from the radar coordinate to \( {P}_r^c = (X_r^c, Y_r^c, Z_r^c)\) in the camera coordinate using \cref{eq:Pc}. The projection formula in \cref{eq:mulintrinsic} is then used to generate mis-calibrated FV and BEV information maps. For the FV map, the recorded pixel value is computed as ${I}_R^{FV}(u_f, v_f) = Z_r^c$; as for the BEV map, the recorded value is determined as ${I}_R^{BEV}(u_b, v_b) = y_r^c$, with $y_r^c$ being $Y_r^c$ plus an offset of the camera height (i.e., the distance above the ground) to eliminate negative values. In \cref{eq:mulintrinsic}, \( (u_f, v_f) \) and ($u_b$, $v_b$) are the coordinates of a radar point \( {P}_r\) projected onto FV and BEV planes using projection matrices \( K \) and \( K'\) correspondingly. Here, \( K \) is the original camera intrinsic matrix. The intrinsic parameters are manually pre-defined for \( K'\) based on the map resolution and map center.

\begin{equation}
\label{eq:Pc}
{P}_r^c =
    T_{init} \begin{bmatrix} P_r \\ 1 \end{bmatrix} =
    \begin{bmatrix} R_{init} & t_{init} \\ 0 & 1 \end{bmatrix} \begin{bmatrix} P_r \\ 1 \end{bmatrix},
\end{equation}

\begin{equation}
\label{eq:mulintrinsic}
    \begin{bmatrix} u_f \\ v_f \\ 1 \end{bmatrix} = K\begin{bmatrix} X_r^c/Z_r^c\\Y_r^c/Z_r^c\\1 \end{bmatrix} ,~and\: \:\begin{bmatrix} u_b \\ v_b \\ 1 \end{bmatrix} = K'\begin{bmatrix} X_r^c \\ Z_r^c \\ 1 \end{bmatrix}.
\end{equation}

\subsubsection*{Camera data.} For the camera data, the FV information map ${I}_I^{FV}(u_d,v_d)$ is derived using the Metric Depth Estimation module, which predicts the depth image from the input image. This module employs two sequential methods: DepthAnything\cite{yang2024depth} for relative depth prediction and ZoeDepth\cite{bhat2023zoedepth} for refining it into metric depth, resulting in ${I}_I^{FV}(u_d,v_d)$. From the depth image, we convert each pixel into a pseudo point cloud $P_p = (X_p,Y_p,Z_p)$ based on \cref{eq:inv}, which are then transformed/projected using matrix $K'$ to form the pseudo-BEV image ${I}_I^{BEV}$ similar to the radar case ${I}_R^{BEV}$. 
 \vspace{-0.3cm}

\begin{equation}
\label{eq:inv}
\begin{bmatrix}X_p \quad Y_p \quad Z_p\end{bmatrix}^ \top = K^{-1}\cdot {I}_I^{FV}(u_d,v_d)\cdot\begin{bmatrix}u_d \quad v_d \quad 1\end{bmatrix}^ \top.
\end{equation}
\vspace{-0.5cm}

\subsection{Feature Extraction}
\label{sec:feature extraction}
\begin{figure*}[t]
  \centering
  \begin{subfigure}{0.45\linewidth}
    \vspace{0.4cm}
    \includegraphics[width=\linewidth]{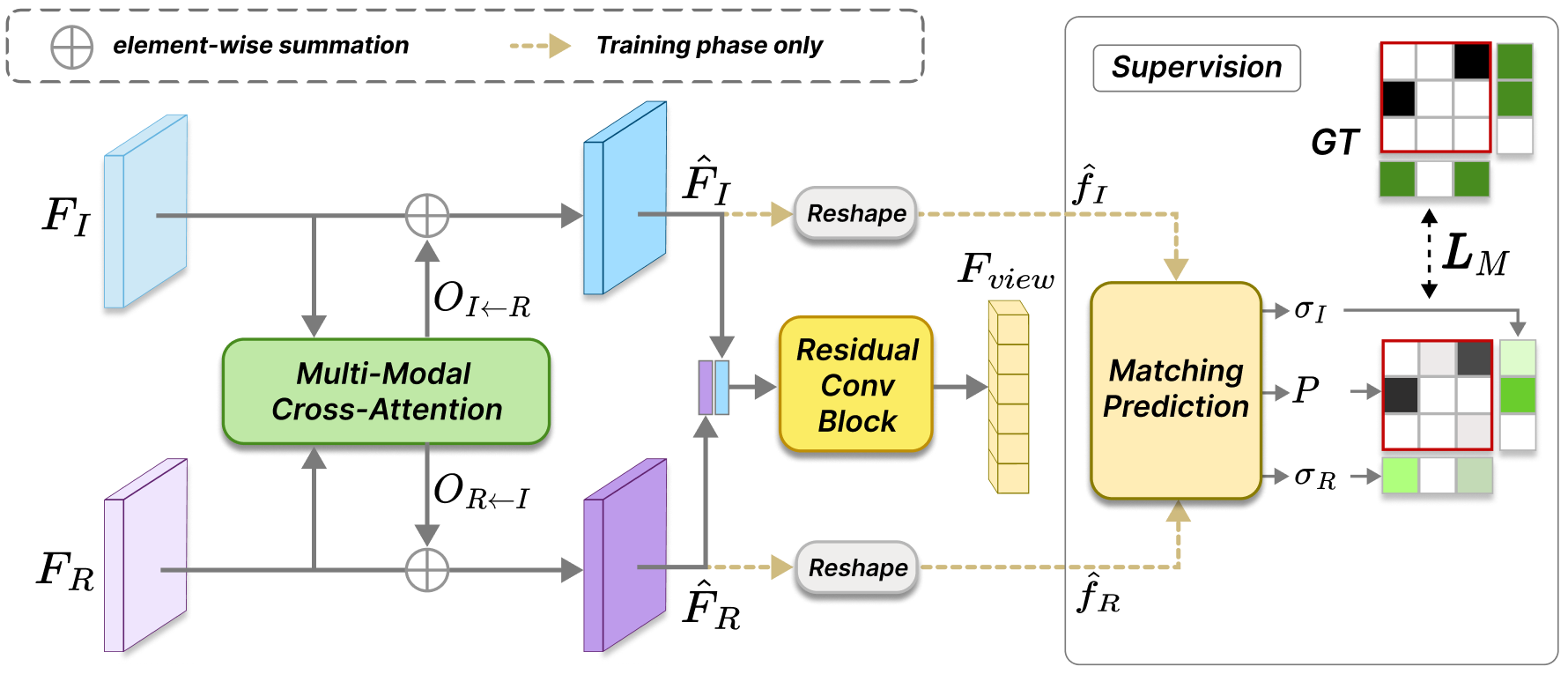}
    \caption{Cross attention and Explicit Feature Matching Supervision}
    \label{fig:feature}
  \end{subfigure}
  \hspace{0.02\linewidth} 
  \begin{subfigure}{0.45\linewidth}
    \includegraphics[width=\linewidth]{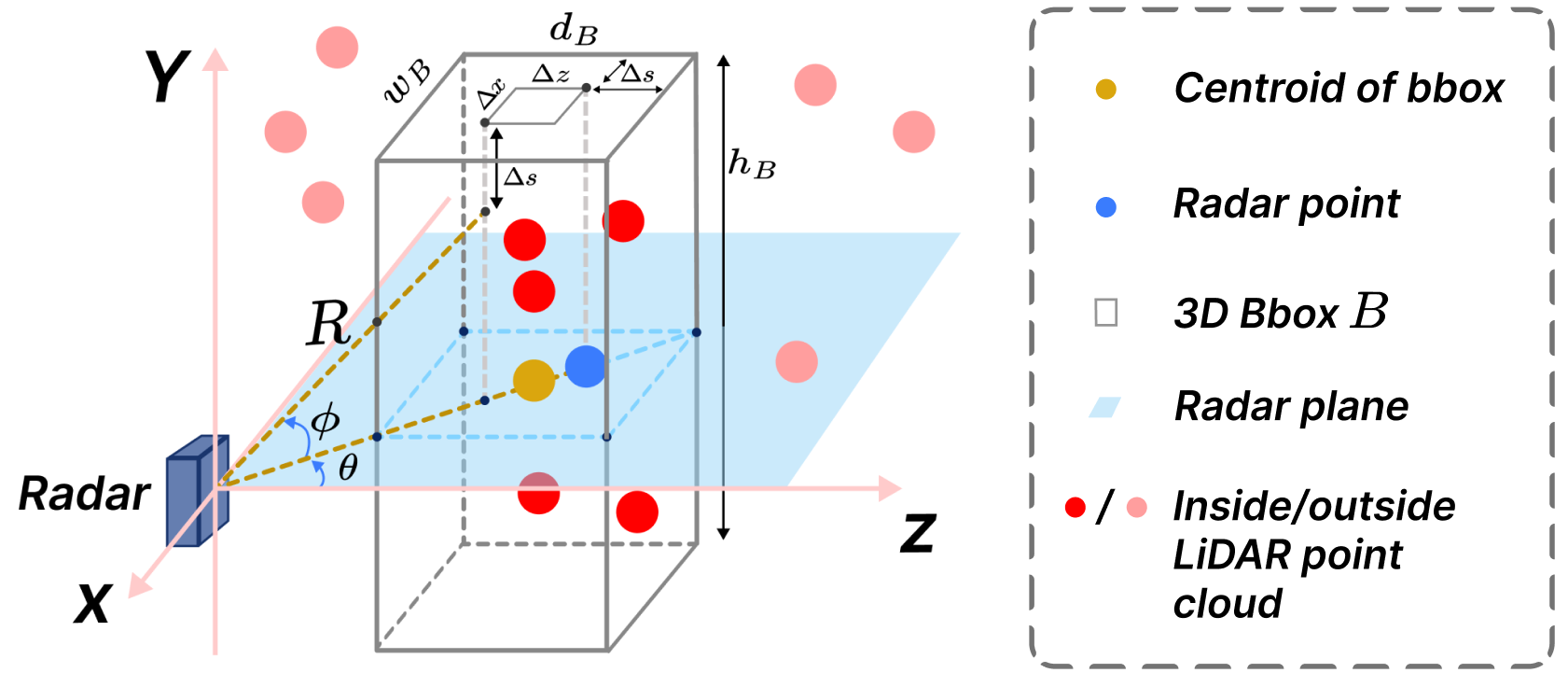}
    \caption{3D bbox $B$ illustration}
    \label{fig:Noise_f}
  \end{subfigure}
  \caption{Illustration of the proposed Feature Matching module.}
  \vspace{-0.3cm}
\end{figure*}
After transforming point clouds and images into two unified representations—FV depth maps (${I}_R^{FV}$, ${I}_I^{FV} \in \mathbb{R}^{H \times W}$) and BEV maps (${I}_I^{BEV}$, ${I}_R^{BEV} \in \mathbb{R}^{H' \times W'}$)—we use ResNet \cite{he2016deep} and convolutional layers to extract features from these maps. Additionally, to enhance the FV semantic content, context features are extracted from the original image using ResNet18 and are integrated with the features from ${I}_I^{FV}$. The resulting feature sets, representing different perspectives for radar and camera, are denoted as $F_R^{FV}$, $F_I^{FV} \in \mathbb{R}^{H/8 \times W/8 \times C}$ and $F_R^{BEV}$, $F_I^{BEV} \in \mathbb{R}^{H'/8 \times W'/8 \times C}$. We detail the network architecture in the supplementary.

\subsection{Feature Matching}
\label{sec:feature matching}

To estimate the 6-DoF (degrees of freedom) extrinsic transformation between radar and camera sensors, our model first focuses on identifying corresponding feature pairs from each sensor's perspective. As illustrated in \cref{fig:overall}, we conduct feature matching across two different perspectives. For each perspective, we implement similar matching modules that include a Multi-Modal Cross-Attention Mechanism and an Explicit Feature Matching Supervision. Additionally, in the FV case, we incorporate a Noise-Resistant Matcher.

\subsubsection*{Multi-Modal Cross-Attention Mechanism.}
Considering both perspectives, the projected point cloud data from radar is sparse with mostly zero values, while camera data is dense with rich depth information and geometric features. We propose a Multi-Modal Cross-Attention (MCA) Mechanism that enables the model to focus on non-zero feature regions and identify correlations between camera and radar features. As described in Equation \cref{eq:mca}, the inputs to the MCA are \( F_I \) and \( F_R \), and the outputs are \( O_{I \leftarrow R} \) and \( O_{R \leftarrow I} \). These outputs are used to compute the updated features \(\hat{F}_{I}\) and \(\hat{F}_{R}\) as shown in \cref{eq:crossfinal}.


\vspace{-0.3cm}
\begin{equation}
\label{eq:mca}
  \begin{aligned}
        (O_{I \leftarrow R}, O_{R \leftarrow I}) &= (\Theta( F_I,m_{I \leftarrow R}), \Theta( F_R,m_{R \leftarrow I})) \\
        &= MCA(F_I, F_R)
    \end{aligned}
\end{equation}


\vspace{-0.2cm}

\begin{equation}
\label{eq:crossfinal}
\hat{F}_{I} = F_I + O_{I \leftarrow R}, \quad \hat{F}_{R} = F_R + O_{R \leftarrow I},
\end{equation}

To obtain \( O_{I \leftarrow R} \) and \( O_{R \leftarrow I} \) in \cref{eq:mca}, we first compute the attended features \( m_{I \leftarrow R} \) and \( m_{R \leftarrow I} \) between the radar and image inside MCA. Inspired by \cite{lindenberger2023lightglue, Hiller2024PerceivingLS}, we use \cref{eq:m1} to compute attended features \( m_{I \leftarrow R} \) and \( m_{R \leftarrow I} \), which rely on the cross-attention score \( a_{IR} \) computed by \cref{eq:score}. 
\vspace{-0.1cm}
\begin{equation}
\label{eq:m1}
m_{R{ \leftarrow}I} = \text{Softmax}(a_{IR}^\top)\:V_I,
m_{I{ \leftarrow}R} = \text{Softmax}(a_{IR})\:V_R,
\end{equation}
\vspace{-0.2cm}
\begin{equation}
\label{eq:score}
a_{IR} = {K_{I}^\top}{K_{R}},
\end{equation}

where $V_*$ and $K_*$ are the value and key, respectively, extracted from the feature $F_*$ through linear projections. Note that, we reshape $F_*$ to $(m \times c)$ dimension before projection and $* \in\mathbb\{I,R\}$. After obtaining these attention maps \( m_{I \leftarrow R} \) and \( m_{R \leftarrow I} \), we apply the cross-modal feature refinement function \( \Theta \) to calculate the \( O_{I \leftarrow R} \) and \( O_{R \leftarrow I} \) by \cref{eq:mca} given \( F_I \) and \( F_R \). The details of \( \Theta \) can be found in the supplementary material.


After obtaining  $\hat{F}_I$ and $\hat{F}_R$, as shown in \cref{fig:feature}, a Residual Conv Block is used to aggregate them into corresponding $F_{view}$ for each perspective branch using \cref{eq:fview}, with
$view \in\mathbb\{BEV,FV\}$.
\vspace{-0.1cm}
\begin{equation}
\label{eq:fview}
\begin{aligned}
F_{view} &= \Phi(\text{conv}(\text{concat}(\hat{F}_{I},\hat{F}_{R})) \\
&+\text{conv}(\text{conv}(\text{concat}(\hat{F}_{I},\hat{F}_{R})))),
\end{aligned}
\end{equation}
\vspace{-0.2cm}

where the first term of \( \Phi \) consists of the concatenated features of \( \hat{F}_I \) and \( \hat{F}_R \) after passing through one convolutional layer, while the second term involves passing through two convolutional layers. These terms are then added together to form our Residual Conv Block. Here, "conv" denotes a block that includes convolutional layers followed by the leakyReLU \cite{xu2015empirical} activation function. The block \( \Phi \) sequentially applies the leakyReLU activation function, flattens the features, and utilizes a multi-layer perceptron (MLP).

\subsubsection*{Explicit Feature Matching Supervision.}
Previously, feature matching was implicitly supervised solely by the final calibration loss to understand overall errors without specifically identifying matched pairs. We find this implicit supervision insufficient and propose directly supervising feature matching using true matching pairs generated from the correct calibration matrix. In other words, we add matching prediction and auxiliary loss during training to enhance the understanding of local feature matching pairs between $\hat{F}_I$ and $\hat{F}_R$.

Motivated by \cite{lindenberger2023lightglue}, an extra branch is designed to perform the task of Local Feature Matching during the training phase as shown in \cref{fig:feature}. Assuming each reshaped feature $\hat{f}_* \in\mathbb{R}^{{m}{\times}{c}}$ from $\hat{F}_*$  includes $m$ (i.e., $H/8 \times W/8$) keypoints, with each keypoint having a feature descriptor of dimension $c$. The assignment matrix $P{\in}[0,1]^{m\times m}$ is estimated based on \cref{eq:P}.

\begin{equation}
\label{eq:P}
P = \sigma_I^\top\:\sigma_R\:\text{Softmax}(S^\top)^\top\:\text{Softmax}(S),
\end{equation}
where $S\in\mathbb{R}^{m\times m}$, calculated by equation \cref{eq:S}, is the similarity score matrix between the features extracted from two sensors after the Multi-Modal Cross-Attention step. Meanwhile, $\sigma_*\:{\in}\:[0,1]^{1\times m}$ is the matchable score of feature points, estimated by equation \cref{eq:sigma}, where $* \in\mathbb\{I,R\}$. A point with a high value of $\sigma$ means it is more likely to have a corresponding point on another map.
\begin{figure}[t]
    \centering
    \includegraphics[width=0.8\linewidth]{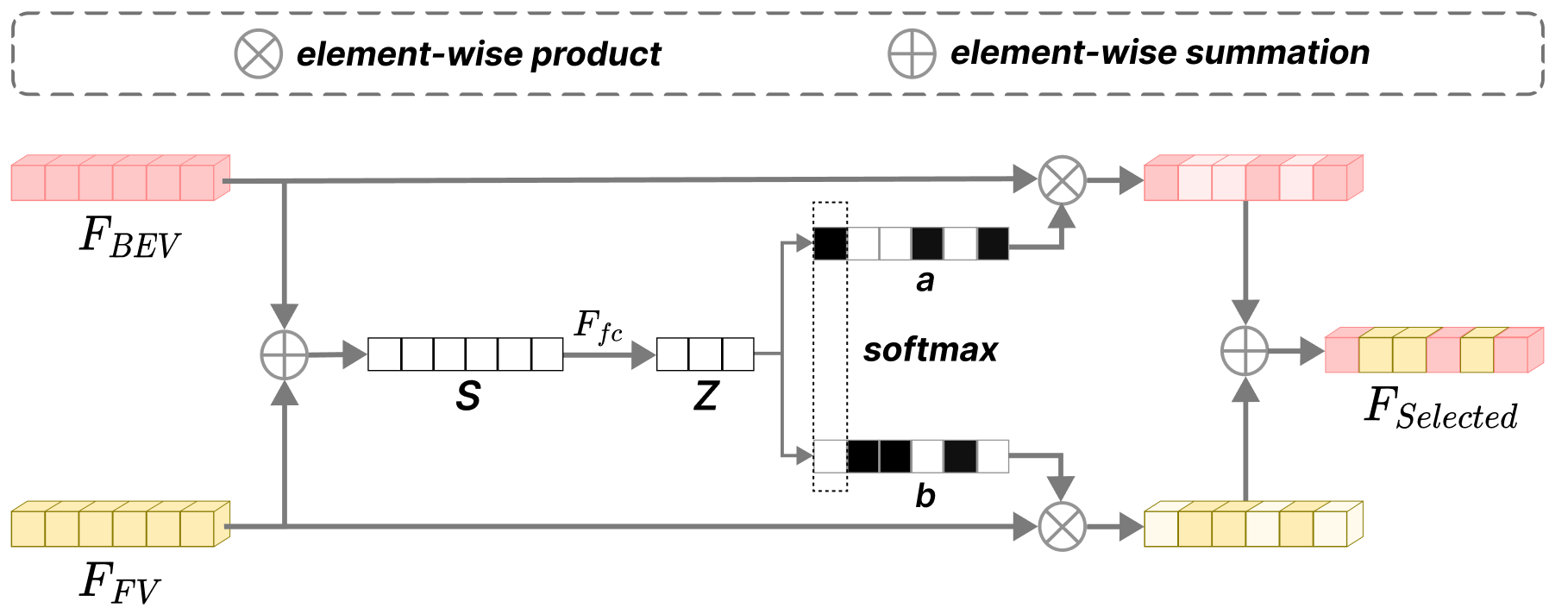}
    \caption{Details of the Selective Fusion Mechanism}
    \vspace{-0.3cm}
    \label{fig:fusion}
\end{figure}
\begin{equation}
\label{eq:S}
S = \text{Linear}(\hat{f}_{I})^\top \text{Linear}(\hat{f}_{R}),
\end{equation}
\vspace{-0.3cm}
\begin{equation}
\label{eq:sigma}
\sigma_* = \text{Sigmoid}(\text{Linear}(\hat{f}_*)).
\end{equation} In the training phase, we use the estimated assignment matrix $P$ and the matching loss defined in \cref{eq:mloss} to directly supervise the cross-sensor matching.

\subsubsection*{Noise-Resistant Matcher.}
\label{sec:noise resistant matcher}
For the FV case, when preparing the ground truth matches matrix $\mathcal{M}$ to supervise the assignment matrix $P$, we recognize that the radar training data contains many unreliable data points due to elevation ambiguity. Additionally, radar signals reflected by objects far from the radar plane can lead to unreliable data points. Therefore, we propose using LiDAR data to identify and remove these unreliable data points from the list of true feature matching pairs $\mathcal{M}$. First, we transform the LiDAR and radar point clouds into a unified camera coordinate system. For each radar 3D point $P_r^c = (X_r^c, Y_r^c, Z_r^c) \in \mathbb{R}^3$, a neighbor region is created using a 3D bounding box $B$. If the number of LiDAR point clouds $P_l^c = (X_l^c, Y_l^c, Z_l^c) \in \mathbb{R}^3$ within this box $B$ exceeds a threshold $\tau$, the radar point cloud $P_r^c$ is considered reliable.

The 3D bounding box $B$ is adaptively designed for each radar point. Denote $\phi$, $\theta$, and $R$ respectively represent the elevation angle, azimuth angle, and point length relative to a 3D radar point. (${\Delta}x$, ${\Delta}z$) are the noises caused by elevation ambiguity as defined in the method \cite{10204897} and ${\Delta}s$ is the error between the two sensors. As shown in \cref{fig:Noise_f}, the box height $h_B$ is calculated by \cref{eq:h} given ${\Delta}s$, $\phi$, $R$, and a predefined parameter $\delta$, which represents the allowable height error from a 3D point to the radar plane.
\vspace{-0.3cm}

\begin{equation}
\label{eq:h}
   h_B = 2({\Delta}y+{\Delta}s) = 2(\delta +{\Delta}s), ~ and  \cos\phi =\sqrt{1-(\delta/R)^2}.
\end{equation}
The width $w_B$ and depth $d_B$ of $B$, are then calculated by \cref{eq:w,eq:d}. The center of a 3D bounding box $B$ of a radar point $P_r^c(X_r^c,Y_r^c,Z_r^c)$ is determined as: $(X_r^c-{\Delta}x/2,\:Y_r^c,\:Z_r^c-{\Delta}z/2)$.
\vspace{-0.4cm}

\begin{equation}
\label{eq:w}
w_B={\Delta}x + 2{\Delta}s = R\sin\theta(1-\cos\phi) + 2{\Delta}s,
\end{equation}
\vspace{-0.3cm}
\begin{equation}
\label{eq:d}
 d_B ={\Delta}z+2{\Delta}s = R\cos\theta(1-\cos\phi) + 2{\Delta}s.
\end{equation}

Due to limited space, we provide additional information and details in the supplementary material.

\subsection{Selective Fusion Mechanism}
\label{sec:selective fusion}

After the Feature Matching step, we can extract the features $F_{BEV}$ and $F_{FV}$ in the corresponding perspective branch. To enhance calibration accuracy by leveraging the distinctive contributions of each perspective, we propose a Dual-Perspective View Selective Fusion Mechanism that combines these features influenced by SKNet \cite{li2019selective}. As illustrated in \cref{fig:fusion}, we first calculate the compact feature $z$ from the sum of the two feature vectors using \cref{eq:z}. Subsequently, a channel-wise attention mechanism adaptively selects diverse elements in the corresponding input features, guided by the compact feature descriptor $z$. The obtained results are then added together to create the final feature $F_{select}$.

\vspace{-0.2cm}

\begin{equation}
\label{eq:z}
z = F_{fc}(F_{BEV}+F_{FV}),
\end{equation}
where $F_{fc}$ denotes the use of a fully connected layer, BatchNorm \cite{ioffe2015batch}, and the leakyReLU activation function.

\subsection{Regression Head}
\label{sec:regression head}
To estimate the rotation and translation parameters and form the updated transformation matrix $T_{pred}$, we leverage the sequence generative decoder from CalibDepth \cite{zhu2023calibdepth}. Specifically, this method employs LSTM \cite{sak2014long}, where the output is defined as a sequence of actions of length $N$ in an autoregressive manner to address the inherent inaccuracies in the one-shot regression approach. After determining $T_{pred}$ at the current iteration, we update $T_{init} = T_{init}\cdot T_{pred}^{i}$, as illustrated in \cref{fig:overall}.
\subsection{Loss Function}
\label{subsec:loss}
Our model operates under two supervised tasks. The primary task focuses on auto-calibration, while an auxiliary task centers on local feature matching, which clarifies the correspondence between feature pairs. Consequently, we employ two corresponding loss functions: the matching loss and calibration loss.

\subsubsection*{Explicit Correspondences Matching.} With ground truth match matrix $\mathcal{M}{\in}\{0,1\}^{N{\times}m{\times}m}$ and the predicted matching $P{\in}[0,1]^{N{\times}m{\times}m}$ and $\sigma_I,\sigma_R \in[0,1]^{N{\times}m}$, the matching loss function is designed to minimize the log-likelihood of the predicted matches as in \cref{eq:mloss}. Since our calibration includes $N$ iterations, the matching loss term aggregates the loss from all iterations.
\vspace{-0.1cm}
\begin{equation}
\label{eq:mloss}
L_M = -\sum_{n}^N\left(   \frac{\lambda}{{\scriptstyle \Sigma{\mathcal{M}}}}L_{pos}^n  + \frac{1-\lambda}{{\scriptstyle \Sigma{\mathcal{N_I}}}+{\scriptstyle \Sigma{\mathcal{N_R}}}} L_{neg}^n\right),
\end{equation}
\vspace{-0.2cm}
\begin{equation}
L_{pos} = \sum_{i,j}  \log(\!P^{ij} )\!\mathcal{M}^{i,j},
\end{equation}
\vspace{-0.2cm}
\begin{equation}
L_{neg} = \sum_{i} \log (1 - \!\sigma_I^{i})\!\mathcal{N}_I^i + \sum_{j}\log (1 - \!\sigma_R^{j})\!\mathcal{N}_R^j,
\end{equation}
\vspace{-0.3cm}

where $\lambda$ is the balancing coefficient between positive and negative instances. $\mathcal{N}_R$ and $\mathcal{N}_I$ are ground truth for ``No Matchable'' scores for pixels in the radar and camera maps respectively. The true matches matrix $\mathcal{M}$ between camera and radar maps is dynamically computed based on the true translation $T_{gt}$ between the two sensors and the current, yet imperfect, updated translation $T_{init}$. $\mathcal{N}_R$ and $\mathcal{N}_I$ are derived from $\mathcal{M}$ by \cref{eq:negative}. 
\begin{equation}
\label{eq:negative}
\mathcal{N}_I^i = 1 - \sum_{j} \mathcal{M}_{ij}, \quad \mathcal{N}_R^j = 1 - \sum_{i} \mathcal{M}_{ij}.
\end{equation}
\vspace{-0.3cm}

The final matching loss function is then computed as the sum of losses from two perspectives defined as $L_{matching} = L_{M_{bev}} + L_{M_{fv}}$.

\subsubsection*{Calibration Loss.}
To both ensure that the calibration results at each iteration step are asymptotic to the ground truth and avoid divergence between different iteration steps, we use the calibration loss $L_{calib}$ proposed in \cite{zhu2023calibdepth}.
By controlling the weight parameter $\beta$, the final total loss consists of the two losses mentioned above, defined as $L_{\text{total}} = L_{\text{calib}} + \beta L_{\text{matching}}$.

\section{Experimental Results}

\subsection{Dataset and Evaluation Metrics}
\subsubsection*{Dataset Preparation.}
We utilize a subset of images from the nuScenes dataset \cite{caesar2020nuscenes} for our training and testing processes. This subset comprises 12,610 samples for training, 1,628 samples for validation, and 1,623 samples for testing. The training and testing depth range spans from 0 to 200 meters, with input and output resolutions set at 400 × 192 pixels.
\subsubsection*{Evaluation Metrics.}
To facilitate comparison with previous work, we convert the output rotation vector to Euler angles. Then, we calculate the absolute error between the predicted values and the ground truth in all dimensions of angles and translation vectors. For all tables, the best and the second-best results are highlighted in bold and underlined, respectively.

\subsection{Main results}
We compared it against the method by Schöller et al. \cite{scholler2019targetless}. Although this method dates back to 2019 and may not incorporate recent advancements, we have also compared it with state-of-the-art LiDAR-camera-based auto-calibration methods, including LCCNet \cite{lv2021lccnet}, CalibDepth \cite{zhu2023calibdepth}, and NetCalib2 \cite{wu2021way}. To ensure a fair comparison, we trained and tested all these methods using the same set of parameters and \textbf{radar-image dataset}. As shown in \cref{tab:compare with SOTA}, our focus lies on testing two mis-calibration ranges: [10°, 0.25m] and [20°, 1.5m], representing small and large error ranges. For ``LCCNet-number'', the ``number'' corresponds to the number of iteration steps.
The achieved results demonstrate that our method exhibits significantly superior average errors in both rotation and translation compared to other methods across both mis-calibration ranges.

\begin{table}[t]
    \centering
    \resizebox{0.475\textwidth}{!}{
    \setlength{\tabcolsep}{2pt}
    \begin{tabular}{ccccccccccc}
    \toprule
    \multirow{2}{*}{Range}&\multirow{2}{*}{ Methods}& \multicolumn{4}{c}{Rotation($^\circ$)} & \multicolumn{4}{c}{Translation(cm)} \\
    &&Mean&Roll&Pitch&Yaw&Mean&X&Y&Z\\
    \midrule
        \multirow{6}{*}{R1}
        
        &LCCNet-1 &1.603&\textbf{0.123}&3.130&1.556&16.531&22.992&17.648&\textbf{8.954}\\
        &NetCalib2 &1.205&0.387&2.289&\textbf{0.941}&\underline{12.297}&\textbf{12.532}&\underline{12.076}&\underline{12.284}\\
        &CalibDepth &\underline{0.807}&0.390&\underline{0.345}&1.686&12.608&12.860&12.250&12.715\\
    \cmidrule(lr){2-10} 
    \arrayrulewidth=0.01mm 
        & Coarse \cite{scholler2019targetless}  &2.035&0.581&1.519&4.004& - & - & - & -\\
        &Fine \cite{scholler2019targetless}   &1.692&0.442&0.939&3.695& - & - & - & -\\&Ours&\textbf{0.427}&\underline{0.130}&\textbf{0.198}&\underline{0.953}&\textbf{9.498}&\underline{12.563}&\textbf{3.295}&12.635\\
    \midrule

        \multirow{7}{*}{R2}
        
        &LCCNet-3&2.156&1.526&2.364&2.579&89.672&\textbf{71.660}&89.605&107.751\\
        &LCCNet-5 &1.898&\underline{0.919}&2.314&2.461&88.302&\underline{74.216}&85.239&105.450\\
        &NetCalib2&2.778&1.465&4.688&\underline{2.180}&71.037&76.001&57.204&79.906\\
        &CalibDepth &\underline{1.686}&1.149&\underline{0.808}&3.102&\underline{55.380}&77.146&\underline{12.918}&\underline{76.078}\\
    \cmidrule(lr){2-10}  
    \arrayrulewidth=0.01mm 
        & Coarse \cite{scholler2019targetless}  &4.388&1.866&3.251&8.048& - & - & - & -\\
        & Fine \cite{scholler2019targetless}  &3.334&1.368&1.937&6.696& - & - & - & -\\          &Ours&\textbf{0.852}&\textbf{0.3597}&\textbf{0.4423}&\textbf{1.7544}&\textbf{47.537}&74.777&\textbf{5.415}&\textbf{62.420}\\
    \bottomrule
    \end{tabular}}
    \vspace{-0.2cm}
    \label{tab:compare with SOTA}
    \caption{Comparison with LiDAR-Camera-Based and Radar-Camera-Based Auto-Calibration Methods on the nuScenes dataset. The methods are compared with two mis-calibration ranges, R1 ($\pm10°, \pm0.25m$) and R2 ($\pm20°, \pm1.5m$).}
\end{table}

\subsection{Ablation Studies}
\begin{table}[t]
    \centering
    \resizebox{0.475\textwidth}{!}{
    \setlength{\tabcolsep}{3pt}
    \begin{tabular}{cccccccccccccc}
    \toprule
        \multirow{2}{*}{FV} & \multirow{2}{*}{BEV} & \multirow{2}{*}{SF}  & \multirow{2}{*}{MCA} & \multirow{2}{*}{EMS} & \multirow{2}{*}{NR} & \multicolumn{4}{c}{Rotation($^\circ$)} & \multicolumn{4}{c}{Translation(cm)} \\
        &&&&&&Mean&Roll&Pitch&Yaw&Mean&X&Y&Z\\
    \midrule
        \ding{51}&&&&&&0.657&0.235& 0.301&1.436&12.602&12.858& 12.247& 12.700\\
        &\ding{51}&&&&&0.689&0.295&0.381&1.392&12.605&12.870&12.285&12.660\\
        \ding{51}&\ding{51}&&&&&0.575&0.209&0.284&1.232&12.315&12.863&11.416&12.667\\
        \ding{51}&\ding{51}&\ding{51}&&&&0.529&0.175&0.237&1.176&11.842&12.882&9.976&12.670\\
        \ding{51}&\ding{51}&\ding{51}&\ding{51}&&&0.502&0.175&0.235&1.097&12.574&12.883&12.150&12.688\\
        \ding{51}&\ding{51}&\ding{51}&\ding{51}&\ding{51}&&\underline{0.463}&\underline{0.140}&\underline{0.206}&\underline{1.042}&\underline{9.627}&\underline{12.563}&\underline{3.682}&\underline{12.636}\\
        \ding{51}&\ding{51}&\ding{51}&\ding{51}&\ding{51}&\ding{51}&\textbf{0.427}&\textbf{0.130}&\textbf{0.198}&\textbf{0.953}&\textbf{9.498}&\textbf{12.563}&\textbf{3.295}&\textbf{12.635}\\
    \bottomrule
    \end{tabular}}
    \vspace{-0.2cm}
    \label{tab:module ablation}
    \caption{Module Impact Ablation (FV: Front View, BEV: Bird's Eye View, SF: Selective Fusion, MCA: Multi-Modal Cross-Attention Mechanism, EMS: Explicit Feature Matching Supervision, NR: Noise-Resistant Matcher)}
\end{table}

\subsubsection*{Impact of each Module.}
\cref{tab:module ablation} shows the impact of each module on experimental outcomes. Using both FV and BEV perspectives together reduced the mean absolute error in rotation by 12.5\%. The Selective Fusion mechanism further reduced this error by 16.5\% by allowing the model to choose suitable features from each perspective. The multi-modal cross-attention mechanism also effectively reduced rotation error by 5\%, demonstrating its capability to address the challenges of sparse radar depth maps. However, the mean absolute error in translation did not improve clearly based on the above modules due to the sparse and noisy radar point clouds. Next, introducing Explicit Feature Matching Supervision decreased the translation error by 23.4\%, highlighting the benefit of explicit matching labels in learning correspondences. Finally, the incorporation of the Noise-Resistant Matcher further filtered out highly inaccurate noise points in the FV perspective and aided in translational calibration. As a result, the mean absolute error in rotation decreased by 7.8\%

\subsubsection*{Dual-Perspective Fusion Methods.}
Additionally, we compare our Selective Fusion Mechanism with commonly used methods such as add fusion and concatenation fusion. In \cref{tab:dual perspective fusion methods}, the results demonstrate a significant improvement of the proposed method over the conventional fusion methods.
\begin{table}[t]
    \centering
    
    \resizebox{0.475\textwidth}{!}{%
    \huge
    \begin{tabular}{ccccccccc}
    \toprule
        \multirow{2}{*}{Fusion Method}& \multicolumn{4}{c}{Rotation($^\circ$)} & \multicolumn{4}{c}{Translation(cm)} \\
        &Mean&Roll&Pitch&Yaw&Mean&X&Y&Z\\
    \midrule
        Add Fusion&0.642&0.217&0.371&1.338&12.547&\textbf{12.832}&12.096&12.714\\
        Concat Fusion&\underline{0.575}&\underline{0.209}&\underline{0.284}&\underline{1.232}&\underline{12.315}&\underline{12.863}&\underline{11.416}&\textbf{12.667}\\
        Selective Fusion&\textbf{0.529}&\textbf{0.175}&\textbf{0.237}&\textbf{1.176}&\textbf{11.842}&12.882&\textbf{9.976}&\underline{12.669}\\
    \bottomrule
    \end{tabular}}
    \vspace{-0.2cm}
    \label{tab:dual perspective fusion methods}
    \caption{Comparison of Dual-Perspective Fusion Methods with Basic Methods}
\end{table}

\subsubsection*{Weight of Explicit Feature Matching Loss.}
In \cref{tab:matching loss ablation}, we test three $\beta$ values: 0.1, 0.3, and 0.5, excluding the noise-resistant matcher module in our method. $\beta$ = 0.5 achieved the lowest rotation error but the highest translation error. $\beta$ = 0.1 provided the best translation accuracy with negligible rotation error. Thus, $\beta$ = 0.1 was selected as the optimal balanced coefficient.
\begin{table}[t]
    \centering
    
    \resizebox{0.475\textwidth}{!}{\begin{tabular}{ccccccccc}
    \toprule
        \multirow{2}{*}{$\beta$}& \multicolumn{4}{c}{Rotation($^\circ$)} & \multicolumn{4}{c}{Translation(cm)} \\
        &Mean&Roll&Pitch&Yaw&Mean&X&Y&Z\\
    \midrule
        0.1&\underline{0.470}&\textbf{0.150}&\underline{0.215}&1.044&\textbf{10.304}&12.587&\textbf{5.684}&\underline{12.640}\\
        0.3&0.471&\underline{0.160}&0.224&\underline{1.029}&\underline{10.377}&\underline{12.555}&\underline{5.940}&\textbf{12.637}\\
        0.5&\textbf{0.446}&0.164&\textbf{0.214}&\textbf{0.960}&12.112&\textbf{12.549}&11.151&\textbf{12.637}\\
    \bottomrule
    \end{tabular}}
    \vspace{-0.2cm}
    \label{tab:matching loss ablation}
    \caption{Ablation Study on Weight Impacts in Explicit Feature Matching Loss}
    \vspace{-0.2cm}

\end{table}

\subsection{Qualitative Results}

\begin{figure}[t!]
  \begin{subfigure}{0.485\linewidth}
      \centering
      \begin{subfigure}[b]{\linewidth}
        \centering
        \includegraphics[width=0.99\linewidth]{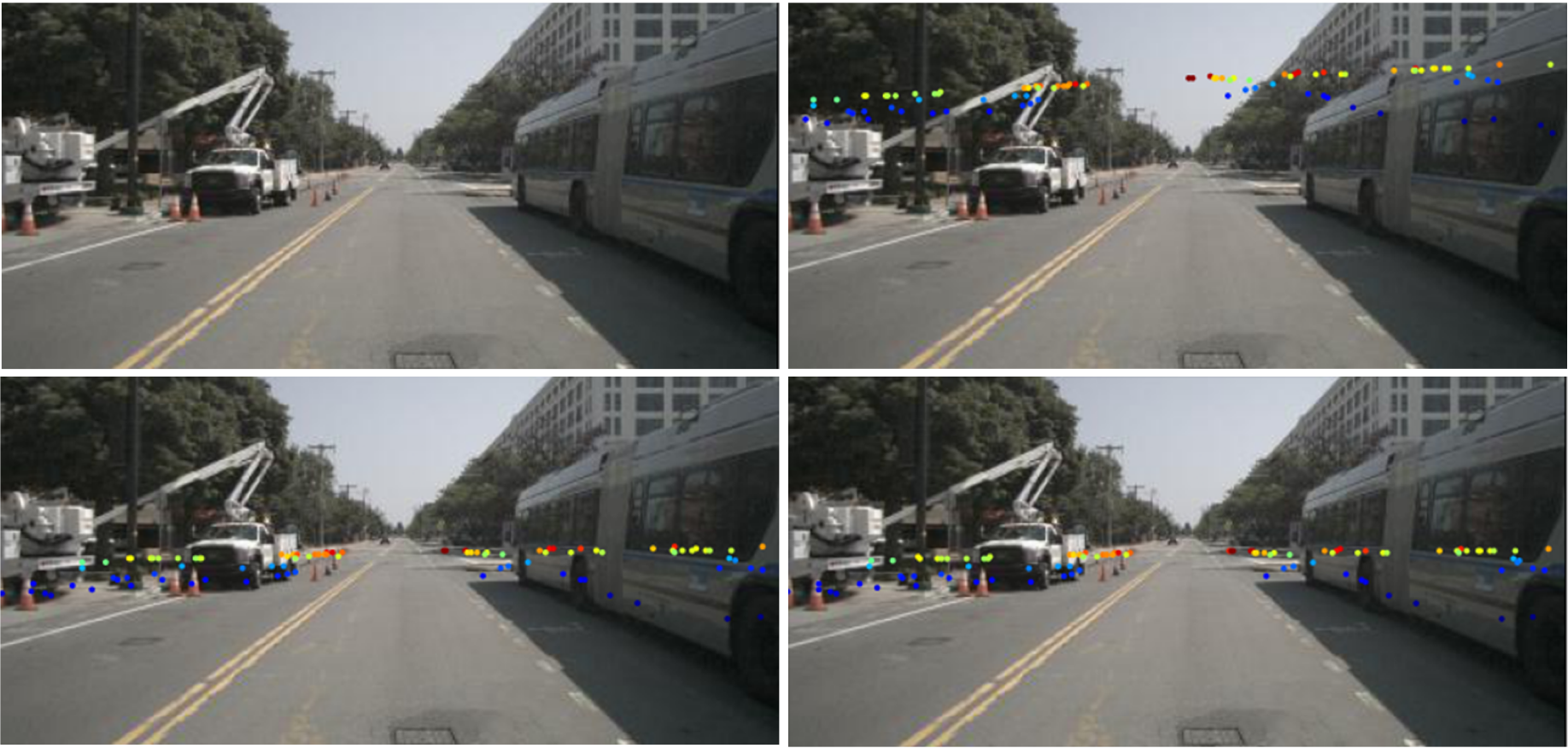}
        \caption*{Sample 1}
      \end{subfigure}
      \begin{subfigure}[b]{\linewidth}
        \centering
        \includegraphics[width=0.99\linewidth]{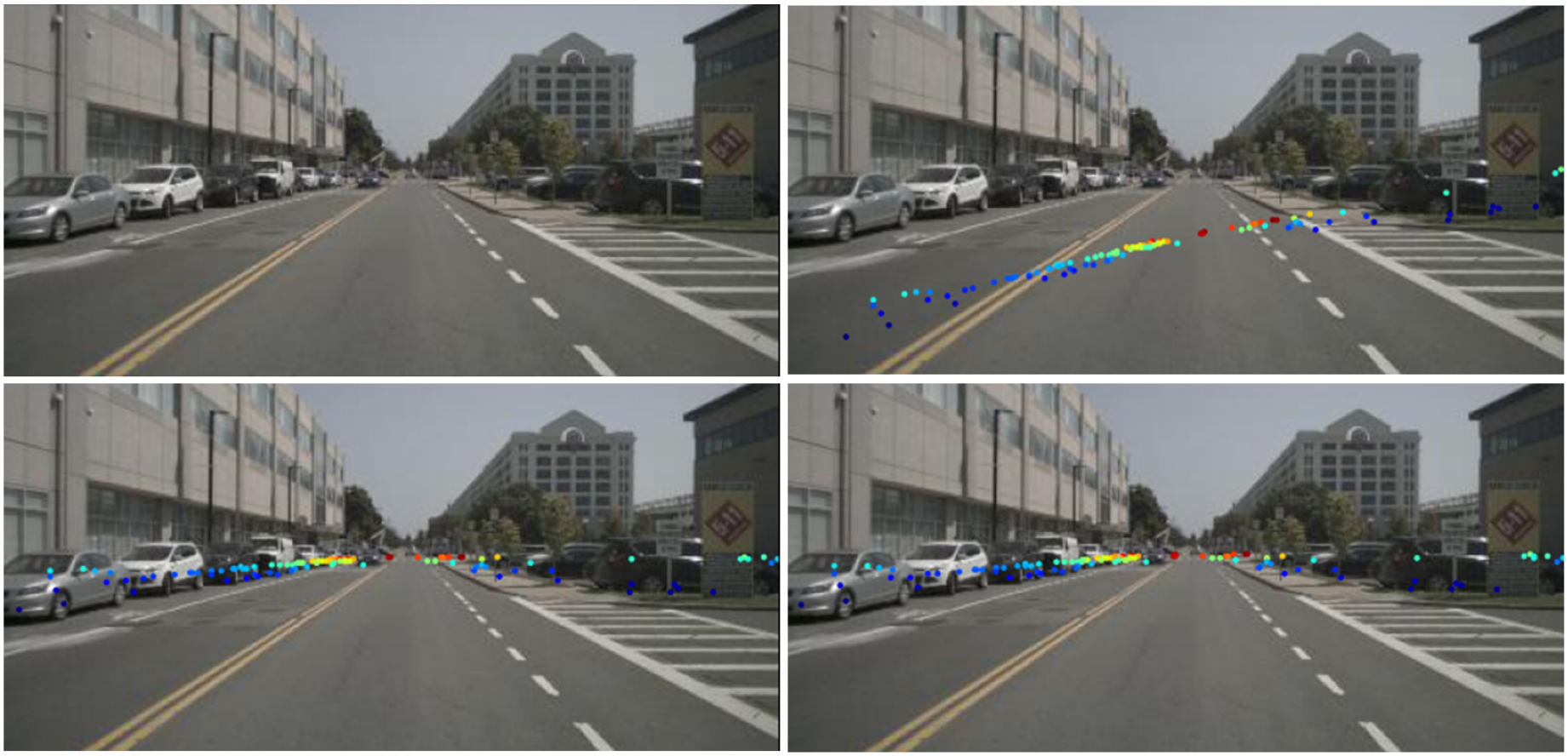}
        \caption*{Sample 2}
      \end{subfigure}
      \caption{Examples of calibration results by projecting radar points onto the frontal-view image. Top-left: input RGB image. Top-right: initial mis-calibrated radar point cloud projection. Bottom-left: network-predicted projection. Bottom-right: ground truth projection.} 
    \vspace{1.65cm}

      \label{fig:visual-a}
  \end{subfigure}
  \hfill
  \begin{subfigure}{0.485\linewidth}
      \centering
      \begin{subfigure}[b]{\linewidth}
        \centering
        \includegraphics[width=0.99\linewidth]{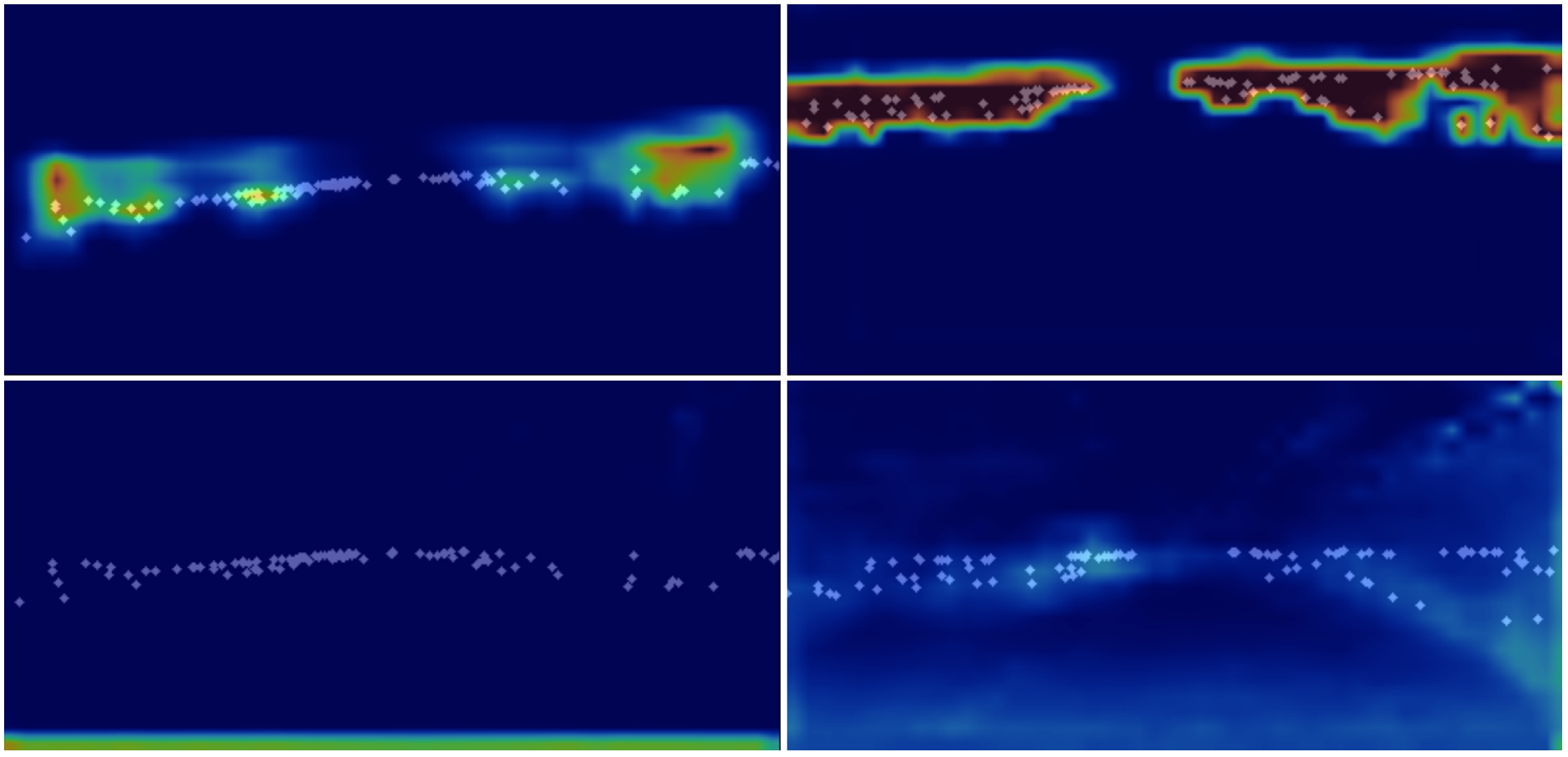}
        \caption*{Sample 1}
      \end{subfigure}
      \begin{subfigure}[b]{\linewidth}
        \centering
        \includegraphics[width=0.99\linewidth]{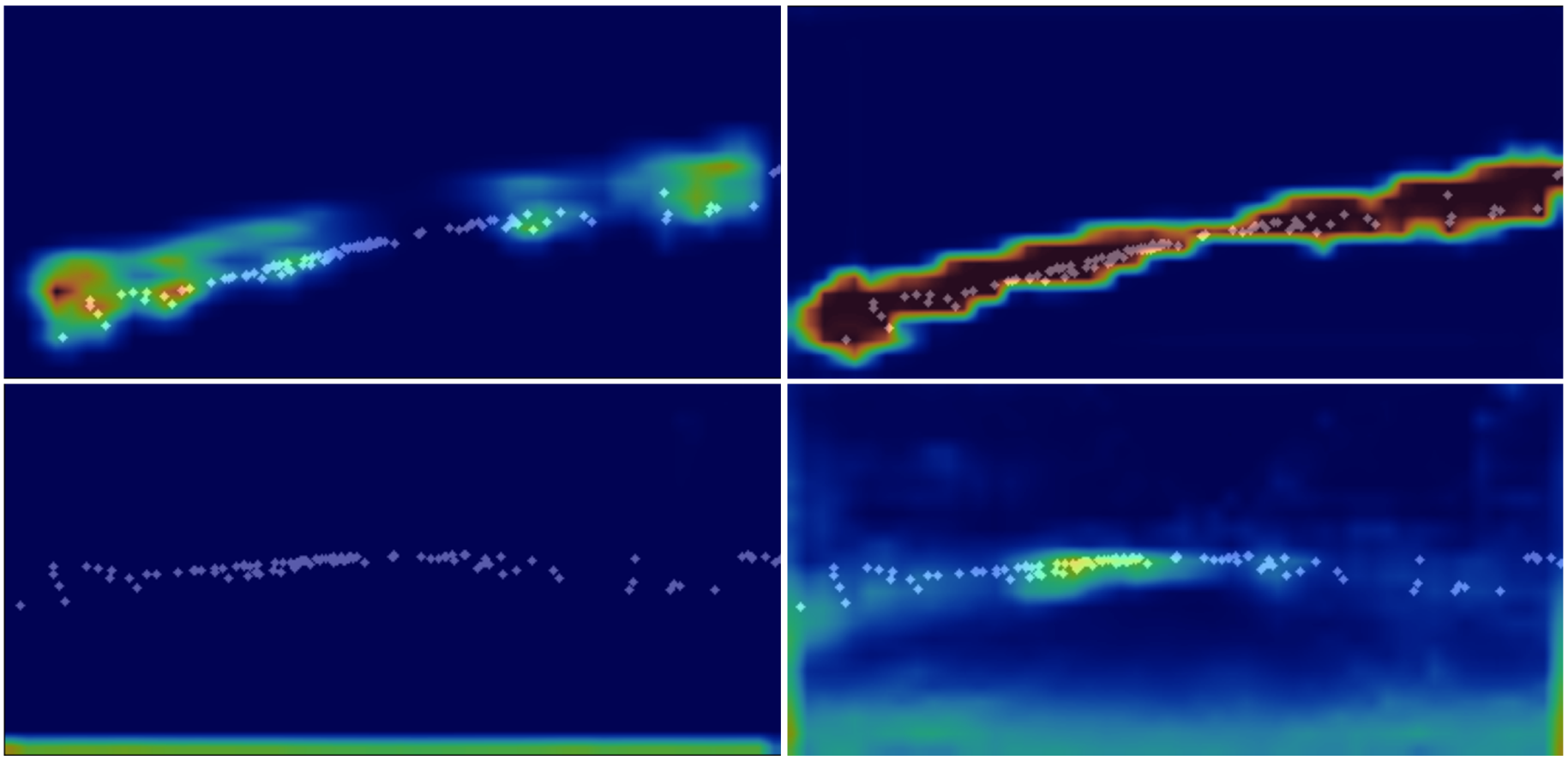}
        \caption*{Sample 2}
      \end{subfigure}
      \caption{Examples of cross-attention maps highlighting the important regions the model focuses on. Top: image-attentive radar importance regions. Bottom: radar-attentive image importance regions. Left/Right: Results with and without explicit matching supervision. The projected 3D points are used to highlight the critical regions. With explicit matching supervision, our model can better identify these critical regions.} 
      
      \label{fig:visual-b}
  \end{subfigure}
  \caption{Results visualization in the front view.}
  \vspace{-0.4cm}
  \label{fig:visual}
\end{figure}

Our method provides accurate calibration results across various initial mis-calibration conditions and scenes. \cref{fig:visual-a} visualizes these results, showing precise calibration even with significant initial errors and sparse radar points. To highlight the effectiveness of our explicit matching supervision, we compute the cross-attention maps $I_{I\leftarrow R}$, $I_{R\leftarrow I}$ between radar and image features, as visualized in \cref{fig:visual-b}. The detailed formulas for computing these attention maps are provided in the supplementary material. We represent the attention maps using heatmaps and denote the projected radar points with white color to indicate critical regions. Without explicit matching supervision, we observe that due to many zero-value relationships, the image-attentive radar exhibits weaker and less focused attention, while the radar-attentive image focuses on non-critical region, such as the ground. After incorporating explicit matching supervision, the image-attentive radar's attention becomes more concentrated on the non-zero regions, specifically the areas of radar point cloud projection. Meanwhile, the radar-attentive image effectively focuses on crucial regions, such as radar projection locations and vehicle contours, rather than being restricted to non-critical region.

\section{Conclusion}
In this work, we present RC-AutoCalib, an end-to-end network for 3D radar and camera calibration. By incorporating dual perspectives, we address the elevation ambiguity in 3D radar. Our Selective Fusion Mechanism integrates useful features from both FV and BEV perspectives. We also developed a Feature Matching module with a Multi-Modal Cross-Attention Mechanism to enhance radar point cloud utilization and a Noise-Resistant Matcher to filter out height-inaccurate noise. Our method achieves a calibration error of 0.427° in rotation and 9.498 cm in translation on the nuScenes dataset, demonstrating competitive performance with these SOTA auto-calibration methods using dense point clouds and establishing a benchmark for future research in 3D radar and camera calibration.

\subsubsection*{Acknowledgments}
\small {This work was financially supported in part (project number: 112UA10019) by the Co-creation Platform of the Industry Academia Innovation School, NYCU, under the framework of the National Key Fields Industry-University Cooperation and Skilled Personnel Training Act, from the Ministry of Education (MOE) and industry partners in Taiwan.  It also supported in part by the National Science and Technology Council, Taiwan, under Grant NSTC-112-2221-E-A49-089-MY3, Grant NSTC-110-2221-E-A49-066-MY3, Grant NSTC-111- 2634-F-A49-010, Grant NSTC-112-2425-H-A49-001, and in part by the Higher Education Sprout Project of the National Yang Ming Chiao Tung University and the Ministry of Education (MOE), Taiwan. We also would like to express our gratitude for the support from MediaTek Inc, Hon Hai Research Institute (HHRI),  E.SUN Financial Holding Co Ltd, Advantech Co Ltd, Industrial Technology Research Institute (ITRI).}

\clearpage

\clearpage
\setcounter{page}{1}
\setcounter{table}{0}
\setcounter{figure}{0}
\setcounter{equation}{0}

\setcounter{section}{0}  

\maketitlesupplementary

\renewcommand{\thesection}{\Alph{section}}

\section{Detail of Feature Extraction}
First, we transform point clouds and images into two unified representations: frontal view depth maps ($I_R^{FV}$, $I_I^{FV} \in \mathbb{R}^{H \times W}$) and BEV images ($I_I^{BEV}$, $I_R^{BEV} \in \mathbb{R}^{H' \times W'}$). These representations allow us to effectively compare and fuse sensor data from different perspectives.

To extract features from radar data (the mis-calibrated images $I_R^{FV}$, $I_R^{BEV}$), we employ the first three blocks of ResNet \cite{he2016deep} as the network structure. This setup, which includes convolutional and pooling layers, is well-suited for extracting low-level image features such as edges and textures. Additionally, considering the sparse nature of radar data and its distinct characteristics compared to image data, we train the radar-specific ResNet from scratch to effectively capture the relevant features.

For the depth map $I_I^{FV}$ and pseudo-BEV map $I_I^{BEV}$, which are derived from the input image using a depth estimation network, feature extraction is performed using just two convolutional layers. Given that these image-derived maps are rich in semantic information, this simplified network configuration has proven sufficient for extracting detailed features while avoiding unnecessary complexity

To enhance semantic content in the frontal view, context features are extracted from the original input image using ResNet18's first three blocks with pretrained weights from ImageNet \cite{deng2009imagenet}. These blocks excel in capturing rich contextual information, which is integrated with features extracted from depth map $I_I^{FV}$ to produce a comprehensive feature representation enriched with semantic information. This fusion not only enhances semantic details in the frontal view but also improves contrast and consistency across multi-view features.

Finally, we obtain feature sets from different perspectives for radar and camera, represented as \( F_R^{FV} \), \( F_I^{FV} \in \mathbb{R}^{H/8 \times W/8 \times C} \) and \( F_R^{BEV} \), \( F_I^{BEV} \in \mathbb{R}^{H'/8 \times W'/8 \times C} \), where \( H \) and \( W \) are the dimensions of the frontal view image, and \( H' \) and \( W' \) are the dimensions of the BEV image.

\begin{figure}[t]
    \centering
    \includegraphics[width=0.8\linewidth]{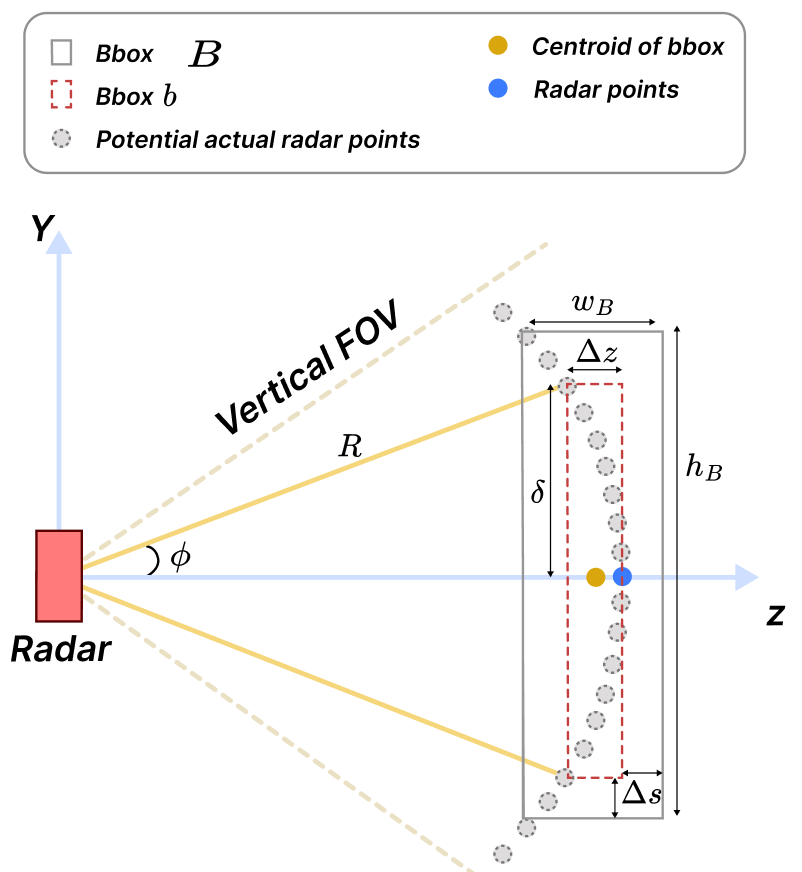}
    \caption{Illustration of bounding box $B$. Suppose we consider only the y and z axes to calculate 
$w_B$ based on $\delta$}
    \label{fig:noise}
\end{figure}
\section{Detail of Multi-Modal Cross-Attention Mechanism}
The output \( O_{I \leftarrow R} \) of the Multi-Modal Cross-Attention Mechanism, as shown in \cref{eq:o_i}, is computed by concatenating the image feature \( f_I \), reshaped from \( F_I \) to dimensions \((m \times c)\), with the attended feature \( m_{I \leftarrow R} \). This concatenated feature is then processed through a feed-forward network (FFN) that employs LayerNorm \cite{ba2016layer}, GELU \cite{hendrycks2016gaussian} activation functions, and linear layers, resulting in the output reshaped to \( O_{I \leftarrow R} \in \mathbb{R}^{h \times w \times c} \). Similarly, \( O_{R \leftarrow I} \) is computed using the same process, as shown in \cref{eq:o_r}.

\vspace{-0.1cm}

\begin{equation}
\label{eq:o_i}
\begin{aligned}
O_{I{ \leftarrow}R} &= \Theta( F_I,m_{I{ \leftarrow}R}) \\
&= \text{reshape}(\text{FFN}(concat[ f_I,m_{I{ \leftarrow}R}]),(h,w,c)),
\end{aligned}
\end{equation}
\begin{equation}
\label{eq:o_r}
\begin{aligned}
O_{R{ \leftarrow}I} &= \Theta( F_R,m_{R{ \leftarrow}I}) \\
&= \text{reshape}(\text{FFN}(concat[ f_R,m_{R{ \leftarrow}I}]),(h,w,c)),
\end{aligned}
\end{equation}

\vspace{0.4cm}

The cross-attention maps $I_{I\leftarrow R}$, $I_{R\leftarrow I}$ between radar and image features will be computed according to the following equation:

\begin{equation}
\label{eq:visual}
I_{I\leftarrow R} = \text{reshape}( \max_j^m(\text{Softmax}(a_{IR})_{ij}),(h,w,1))
\end{equation}
\vspace{-0.2cm}
\begin{equation}
\label{eq:visual2}
I_{R\leftarrow I}= \text{reshape}( \max_j^m(\text{Softmax}(a_{IR}^\top)_{ij}),(h,w,1))
\end{equation}

\section{Detail of Noise-Resistant Matcher}
\begin{figure*}[t]
    \centering
    \includegraphics[width=0.95\linewidth]{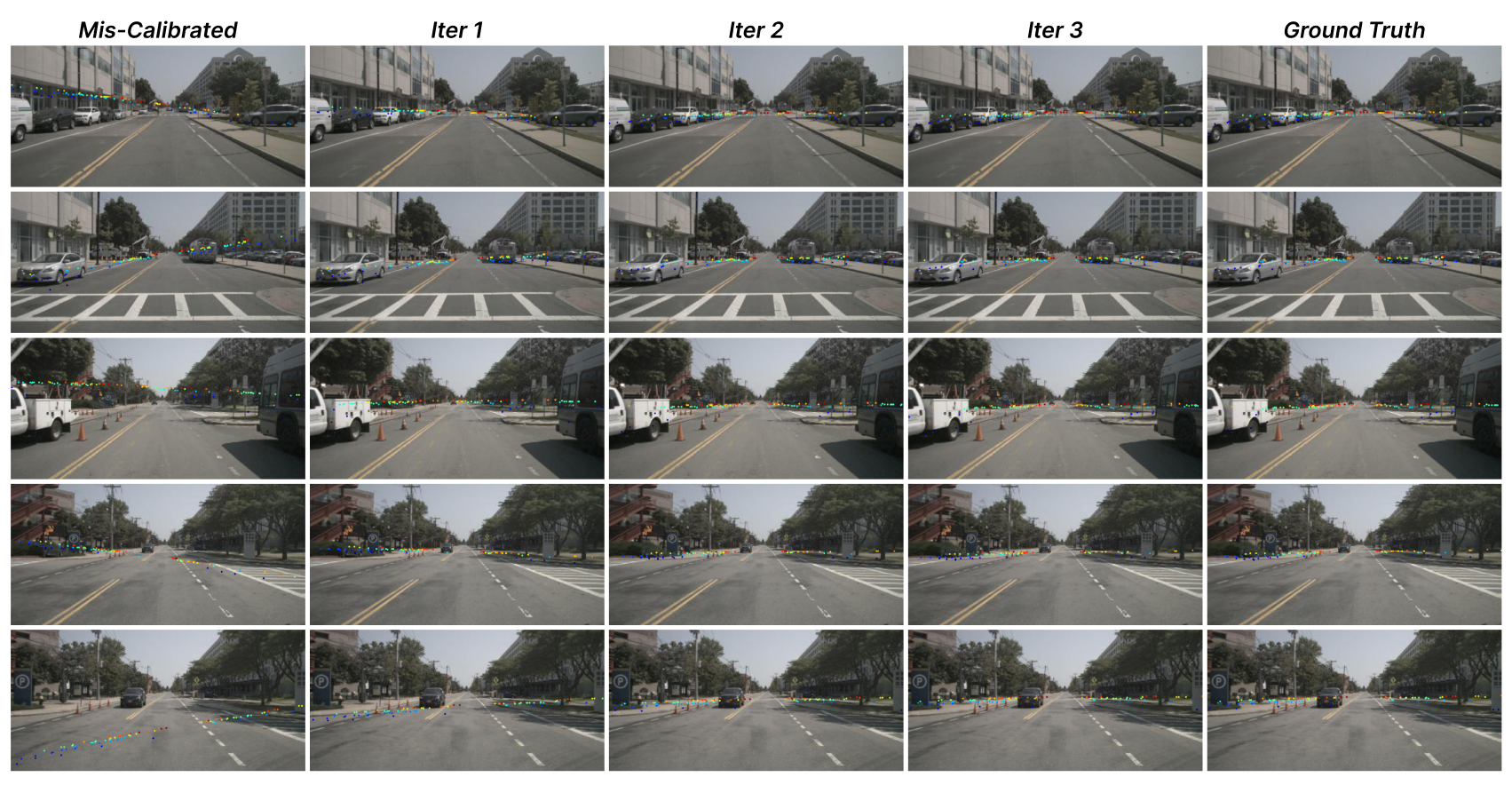}
    \caption{Calibration results by projecting
radar points onto the FV image}
    \label{fig:visualizesupp}
\end{figure*}

\begin{figure*}[t]
    \centering
    \includegraphics[width=0.95\linewidth]{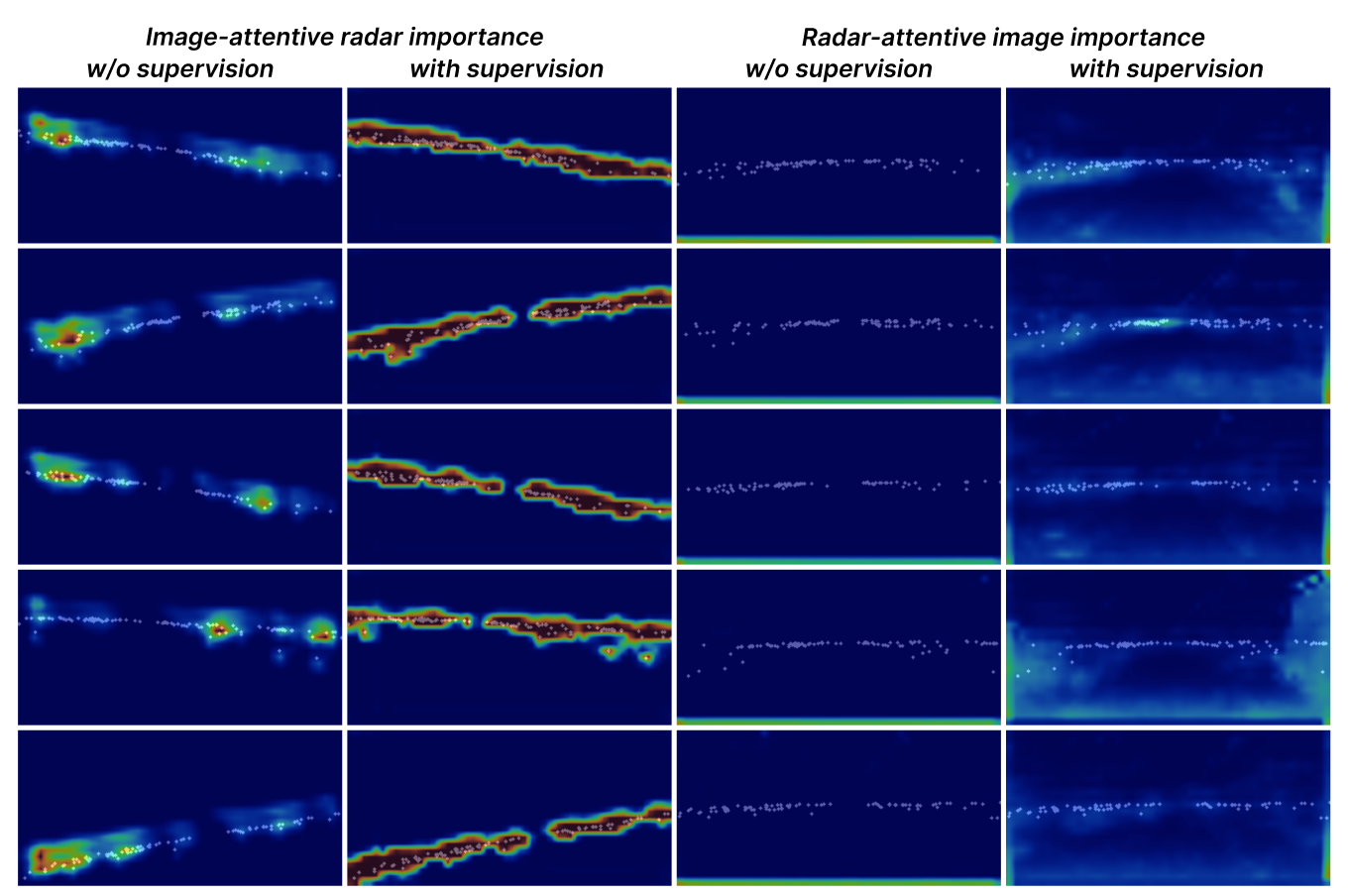}
    \caption{Examples of FV cross-attention maps highlight the important regions the model focuses on.}
    \label{fig:fv attention}
\end{figure*}
\cref{fig:noise} illustrates the principle of the noise-resistant matcher, which simplifies by removing the x-axis. The radar point cloud is computed based on azimuth angle $\theta$ and distance $R$, all lying on the radar point plane with a constant y-axis value of 0. However, in reality, radar points are reflected from objects at a distance from the radar plane, leading to the appearance of uncertain elevation angle $\phi$ within the vertical FOV boundary. Therefore, we depict the gray points in figure as potential actual radar points within the FOV, at the same distance $R$ but with varying elevation angles $\phi$.

For potential actual radar point, there is an error in both the x and z axes, corresponding to $\Delta x$ and $\Delta z$ as defined in \cite{10204897}. Using these errors, we define a region encompassing neighboring LiDAR points. Essentially, each radar point creates a bounding box $b$ to identify LiDAR neighbors associated with potential actual radar points. This 3D bounding box is fixed with a parameter $\delta$, which is the allowable height error threshold, and the width and depth correspond to $\Delta x$ and $\Delta z$, respectively. In \cref{fig:noise}, when the allowable height limit for the potential actual radar point is $\delta$, the maximum allowable z value for the gray point is when it coincides with the radar point (blue) $P_r^c(X_r^c,Y_r^c,Z_r^c)$, and the minimum allowable z value is at $(Z_r^c- \Delta z)$, similarly for the x-axis. Therefore, the center of the 3D bounding box $b$ is defined as $(X_r^c - \Delta x/2,y_r^c,Z_r^c - \Delta z/2)$.

Additionally, in reality, LiDAR points will not fit exactly with potential actual radar points due to measurement inaccuracies of both the radar and LiDAR sensors. Therefore, we add an offset to the width, height, and depth by a fixed error $\Delta s$, forming the 3D bounding box $B$. Both parameters $\delta$ and $\Delta s$ are tuned based on the unit meter.




\section{Implementation Details}
\begin{table}[t]
    \centering
    \resizebox{0.475\textwidth}{!}{
    \begin{tabular}{ccccccccccc}
    \toprule
    \multirow{2}{*}{Scenario} & \multirow{2}{*}{Methods} & \multicolumn{3}{c}{Rotation ($^\circ$)} \\
    \cmidrule(lr){3-5}
    && Mean & Roll & Pitch & Yaw \\
    \midrule
    \multirow{6}{*}{Urban} 
        & LCCNet-1 & 2.969 & 3.123 & \underline{2.703} & \underline{3.081} \\
        & NetCalib2 & \underline{2.643} & \textbf{0.742} & 3.221 & 3.966 \\
        & CalibDepth & 3.656 & 2.088 & 3.913 & 4.966 \\
        & Coarse~\cite{scholler2019targetless} & 4.395 & 3.148 & 4.645 & 5.392 \\
        & Fine~\cite{scholler2019targetless} & 4.956 & 3.152 & 5.195 & 6.520 \\
        & Ours & \textbf{1.875} & \underline{0.934} & \textbf{2.609} & \textbf{2.082} \\
    \midrule
    \multirow{6}{*}{Rain} 
        & LCCNet-1 & \underline{2.394} & 3.400 & \textbf{1.929} & \textbf{1.853} \\
        & NetCalib2 & 2.612 & \underline{0.644} & \underline{2.781} &4.411 \\
        & CalibDepth & 3.412 & 1.686 & 4.299 & 4.251 \\
        & Coarse~\cite{scholler2019targetless} & 4.092 & 2.060 & 4.663 & 5.554 \\
        & Fine~\cite{scholler2019targetless} & 4.776 & 2.050 & 5.269 & 7.007 \\
        & Ours & \textbf{1.922} & \textbf{0.622} & 3.273 & \underline{1.870} \\
    \bottomrule
    \end{tabular}}
    \caption{Cross-dataset evaluation on the aiMotive dataset. 
    The nuScenes-trained models are evaluated on aiMotive scenarios 
    (urban and rain) with an initial rotation error range within 10\textdegree.}
    \label{tab:compare_cross}
\end{table}

We resized the original 1600x900 images to 400x192 pixels. Training was conducted on an NVIDIA GTX 3090 GPU for 50 epochs using the Adam optimizer with an initial learning rate of 1e-4, halving it every 8 epochs. The loss function weights were set to $\lambda = 0.75$ and $\beta = 0.1$. In the Regression Head, the LSTM module had a fixed iterative step size of 3.
In the noise-resistant matcher section, we selected a threshold $\tau$ of 3, $\Delta s$ of 0.5, and $\delta$ of 1.

\section{Additional Experimental  Results}

\subsection{Cross-dataset evaluation}
We compared our RC-AutoCalib with other related methods on the \textbf{aiMotive} \cite{matuszka2022aimotive} dataset, as shown in \cref{tab:compare_cross}. In this experiment, all models were trained on the nuScenes dataset and directly tested on two scenarios from \textbf{aiMotive}. Our method outperformed others in both scenarios, demonstrating the superior gener alization ability of our model.

\subsection{Positive-Negative Balance in Feature Matching Supervision Loss}
In \cref{tab:positive negative ablation}, we experimented with different values of $\lambda$ for $L_{matching}$, including 0.9, 0.75, and 0.5. When $\lambda$ was set to 0.75, both rotation error and translation error reached their lowest values.
\subsection{LiDAR-Camera Calibration}
To showcase the adaptability of our approach, we extended it to LiDAR-camera calibration. We trained our method on the nuScenes dataset using the same train-test split as reported in the main paper and on the KITTI dataset with 24,000 training samples and 6,000 test samples. These experiments were conducted without the Noise-Resistant Matcher, which is specific to radar data.

As shown in \cref{tab:lidar-camera_compare}, we compare our method with previous approaches, including CalibNet\cite{iyer2018calibnet}, CalibRCNN\cite{shi2020calibrcnn}, CalibDNN\cite{zhao2021calibdnn}, CalNet\cite{9956145}, and CalibDepth\cite{zhu2023calibdepth}. The results demonstrate that our method outperforms them, confirming its scalability and robustness.
\subsection{Effects on Downstream tasks}
To validate the impact of our method on 3D object detection, we initialized random incorrect extrinsic parameters, corrected the parameters for each image in the scenes test set, and evaluated the pre-trained CRN \cite{kim2023crn} 3D object detection model. The mAP performance decreased by only \textbf{0.27\%} compared to using ground-truth calibration, indicating a negligible difference.
\begin{table}[t]
    \centering
    
     \resizebox{0.475\textwidth}{!}{
    \begin{tabular}{ccccccccc}
    \toprule
        \multirow{2}{*}{$\lambda$}& \multicolumn{4}{c}{Rotation($^\circ$)} & \multicolumn{4}{c}{Translation(cm)} \\
        &Mean&Roll&Pitch&Yaw&Mean&X&Y&Z\\
    \midrule
        0.9&0.460&\underline{0.142}&0.222&1.017&\underline{10.896}&\underline{12.561}&\underline{7.503}&\textbf{12.625}\\
        0.75&\textbf{0.427}&\textbf{0.130}&\textbf{0.199}&\textbf{0.953}&\textbf{9.498}&12.564&\textbf{3.295}&\underline{12.635}\\
        0.5&\underline{0.442}&0.153&\underline{0.209}&\underline{0.9634}&11.295&\textbf{12.547}&8.699&12.638\\
    \bottomrule
    \end{tabular}}
    \caption{Ablation Study on Positive-Negative Balance in Feature Matching Supervision Loss}
    \label{tab:positive negative ablation}
\end{table}
\begin{table}[t]
    \centering
    
    \resizebox{0.475\textwidth}{!}{
    \setlength{\tabcolsep}{2pt}
    \begin{tabular}{ccccccccccc}
    \toprule
    \multirow{2}{*}{Dataset} & \multirow{2}{*}{Methods} & \multicolumn{4}{c}{Rotation($^\circ$)} & \multicolumn{4}{c}{Translation (cm)} \\
    &&Mean & Roll & Pitch & Yaw & Mean & X & Y & Z \\
    \midrule
        \multirow{6}{*}{KITTI}
        & CalibNet    & 0.410 & 0.150 & 0.900 & 0.181 & 7.82 & 12.10 & 3.49 & 7.87 \\
        & CalibRCNN   & 0.428 & 0.199 & 0.640 & 0.446 & 5.30 & 6.20  & 4.30 & 5.40 \\
        & CalibDNN    & 0.210 & 0.110 & 0.350 & 0.180 & 5.07 & 3.80  & 1.80 & 9.60 \\
        & CalNet      & 0.200 & 0.100 & 0.380 & \textbf{0.120} & 3.03 & 3.65  & 1.63 & 3.80 \\
        & Ours        & \textbf{0.142} & \textbf{0.066} & \textbf{0.096} & 0.268 & \textbf{1.941} & \textbf{2.479} & \textbf{0.998} & \textbf{2.347} \\
    \midrule
        \multirow{2}{*}{nuScenes}
        & CalibDepth  & 0.408 & 0.215 & 0.226 & 0.794 & 8.33  & 11.19 & 4.27  & 9.53  \\
        & Ours        & \textbf{0.208} & \textbf{0.142} & \textbf{0.148} & \textbf{0.337} & \textbf{3.183} & \textbf{1.010} & \textbf{0.7836} & \textbf{7.836} \\
    \bottomrule
    \end{tabular}}
    \vspace{-0.1cm}
    \caption{Comparison of the method extension to the LiDAR-Camera auto-calibration task on the nuScenes and KITTI datasets. The methods are compared with mis-calibration ranges R1 ($\pm10^\circ$, $\pm0.25m$). Notably, the CalibDepth method was retrained on the nuScenes dataset by us.}
    \label{tab:lidar-camera_compare}
\end{table}

\subsection{Qualitative Results}
\cref{fig:visualizesupp} shows additional calibration results, including the results for each iteration. It can be observed that even with a large initial error, our method effectively reduces the error progressively with each iteration. In \cref{fig:fv attention}, we present additional attention maps using heatmaps in the FV, with the projected radar points marked in white to indicate critical regions. 
\vspace{1cm}


{
    \small
    \bibliographystyle{ieeenat_fullname}
    \bibliography{main}
}


\end{document}